\documentclass[a4paper,twoside]{article}

\usepackage{epsfig}
\usepackage{subcaption}
\usepackage{calc}
\usepackage{amssymb}
\usepackage{amstext}
\usepackage{amsmath}
\usepackage{amsthm}
\usepackage{multicol}
\usepackage{pslatex}
\usepackage{apalike}
\usepackage{algorithm2e}
\usepackage[bottom]{footmisc}

\usepackage{xurl}
\usepackage{multirow}
\usepackage[hidelinks]{hyperref}
\usepackage{mathtools}
\usepackage{siunitx}
\usepackage[inline]{enumitem}
\usepackage{tikz}
\usetikzlibrary{positioning,calc,arrows.meta}
\usepackage[dvipsnames]{xcolor}

\usepackage{SCITEPRESS}     

\begin{document}

\title{%
Keeping the Franka Emika Panda alive: a ROS~2 stack\\with a reliable position interface
}
    
\author{%
\authorname{Antonio Langella\orcidAuthor{0009-0009-7656-8325},
Davide Risi\orcidAuthor{0009-0001-3077-5371},
Vincenzo Petrone\orcidAuthor{0000-0003-4777-1761},
Enrico Ferrentino\orcidAuthor{0000-0003-0768-8541}, and
Pasquale Chiacchio\orcidAuthor{0000-0003-3385-8866}
}
\affiliation{Department of Information Engineering, Electrical Engineering, and Applied Mathematics (DIEM), University of Salerno, 84084 Fisciano, Italy}
\email{\{alangella, drisi, vipetrone, eferrentino, pchiacchio\}@unisa.it}
This work has been accepted for publication at ICINCO 2026 - 23rd International Conference on Informatics in Control,
Automation and Robotics on Thursday 10th September, 2026.\\
Science and Technology Publications, Lda holds the copyright on the published version of this article.\\
Please refer to the published version in the conference proceedings at \url{https://doi.org/10.5220/0015508000004979}
}


\abstract{%
This paper presents an open-source software stack that restores ROS 2 support for the Franka Emika Panda robot while resolving the long-standing unreliability of its external position control interface.
We first analyze the root causes of unstable position control and show that the observed vibrations and protective stops arise from the timing of the external control loop and sampling jitter, rather than from limitations of the robot itself.
Building on this analysis, we introduce an asynchronous hardware interface that decouples real-time communication from the ROS 2 control loop, a rate-matching mechanism for slower command sources, and a position-domain reference generation strategy that produces reliable, smooth position commands.
Experimental validation shows that the proposed architecture reliably tracks velocity references by reducing motion artifacts introduced by the official implementation, and the stack is validated across motion planning, compliance control, position-controlled manipulation, and haptic teleoperation on two independent Panda platforms.
By restoring a modern, reliable, and open ROS 2 ecosystem for the Panda, this work lowers the barrier to developing safe, responsive, and reproducible human-robot collaboration applications that integrate planning, perception, interaction, and shared autonomy.
Code and videos are available on our website at \url{https://sites.google.com/unisa.it/fer-ros2}.
}

\onecolumn \maketitle \normalsize \setcounter{footnote}{0} \vfill


\section{\uppercase{Introduction}}
\label{sec:introduction}

The Franka Emika Robot (FER), widely known in the community as the Franka Emika Panda (Panda, in the following), is one of the most common platforms for manipulation research and collaborative robotics \cite{haddadin_franka_2022}.
Its torque-controlled arm, force sensitivity and open control interface made it a de-facto standard in academic laboratories, where a large installed base is still in daily use \cite{daniel_multi_2024,de_melo_unified_2025,shahid_gear_2025,petrone_optimized_2026}.
Vendor software support for the Panda, however, has been discontinued: current releases of \texttt{franka\_ros2} and \texttt{libfranka} target the newer Franka Research 3 (FR3) arm and recent ROS~2 distributions \cite{macenski_robot_2022}, but no longer support the Panda \cite{franka_robotics_franka_ros2_2026}.
Laboratories that rely on the Panda are therefore left on outdated, unmaintained software, an obstacle which matters most in collaborative applications, where the robot must be integrated with modern planning, perception and interaction tooling.

A second obstacle concerns position control.
On the Panda, commanding joint positions or Cartesian poses through the external control interface is unreliable: the motion is prone to vibrations and to protective stops, to the point that existing frameworks discourage the position interface altogether \cite{skerlj_multipanda_ros2_2026}.
This is a serious limitation for collaborative use, since position commands are the natural output of, e.g., motion planners or perception-based modules, thus forcing researchers to put additional effort in devising a dedicated control architecture to execute such high-level commands.
As we show in this work, the problem is not inherent to the robot, but to the specific implementation of the external control loop, and it can be resolved.

This work restores support for the Panda on ROS~2 Jazzy, a long-term support distribution, makes its position interface reliable, and releases the code as open source.
Our contributions are:
\begin{enumerate}
  \item An analysis of why external position control performs poorly on the Panda, tracing it to the timing of the control interface rather than to the robot itself (Sec.~\ref{sec:problem});
  \item An asynchronous hardware interface that decouples the \texttt{ros2\_control} loop from the robot's real-time channel, together with a position-domain reference-generation scheme that makes the position interface reliable, and an optional rate-matching stage that enables slower command sources (Sec.~\ref{sec:architecture});
  \item The porting of \texttt{franka\_ros2} \cite{franka_robotics_franka_ros2_2025} to ROS~2 Jazzy for the Panda, built on an extended and packaged \texttt{libfranka} driver which can run on a stock kernel without a real-time patch, both released as citable open-source repositories\footnote{\url{https://github.com/unisa-acg/fer_moveit_config}};
  \item An experimental validation of the proposed reference-generation scheme (Sec.~\ref{sec:validation}) and a demonstration in collaborative application scenarios (Sec.~\ref{sec:applications}) at two experimental sites, namely the Robotics Lab at University of Salerno and IDSIA, USI-SUPSI, hereinafter ``Salerno'' and ``IDSIA'', respectively.
\end{enumerate}

\subsection{Related Works}
\label{sec:intro:related}

The reference software for Franka arms is the vendor \texttt{franka\_ros2} stack, built on \texttt{ros2\_control} \cite{chitta_ros_control_2017}; its current releases require a \texttt{libfranka} version that supports only the FR3, leaving the Panda unsupported.
The closest work to ours is \texttt{multipanda\_ros2} \cite{skerlj_bridging_2026}, a ROS~2 framework for multi-robot control of Franka arms.
It targets real-time \emph{torque} control and multimanual coordination, with a MuJoCo-based simulation \cite{todorov_mujoco_2012} for bridging the sim-to-real gap, and it is implemented for the torque control mode only, explicitly leaving other modes as future work.
At the reference-generation level, \texttt{Ruckig} \cite{berscheid_jerk-limited_2021} provides real-time kinematically-constrained online trajectory generation for arbitrary target states and demonstrates it on the FER.
Our goals are complementary: we focus on a current single-arm stack and, specifically, on the reliable execution of externally generated \emph{position} commands, which neither \texttt{multipanda\_ros2} nor \texttt{Ruckig} address.

\subsection{Paper Structure}
\label{sec:intro:structure}

The remainder of this paper is structured as follows.
Sec.~\ref{sec:problem} analyzes the Franka Control Interface (FCI) and its official ROS~2 implementation, and identifies the two mechanisms that make external position control fail.
Sec.~\ref{sec:architecture} presents the proposed architecture, and Sec.~\ref{sec:validation} validates it against the official one on the same robot and trajectory.
Sec.~\ref{sec:applications} then demonstrates the stack on four applications chosen to span key capabilities for human-robot collaboration: planning and executing motions in a shared workspace, regulating the compliance the arm reacts to physical contact with, manipulating objects
through position commands, and teleoperating the system through a haptic device.
Each exercises a different command interface of the same hardware interface, and the first two are reproduced on two independent Panda instances.
Lastly, Sec.~\ref{sec:conclusions} concludes the paper.


\section{THE FCI AND ITS ROS~2 IMPLEMENTATION}
\label{sec:problem}

\subsection{FCI Communication Protocol}
\label{sec:problem:fci}

\begin{figure*}
  \centering
  \includegraphics[width=\linewidth]{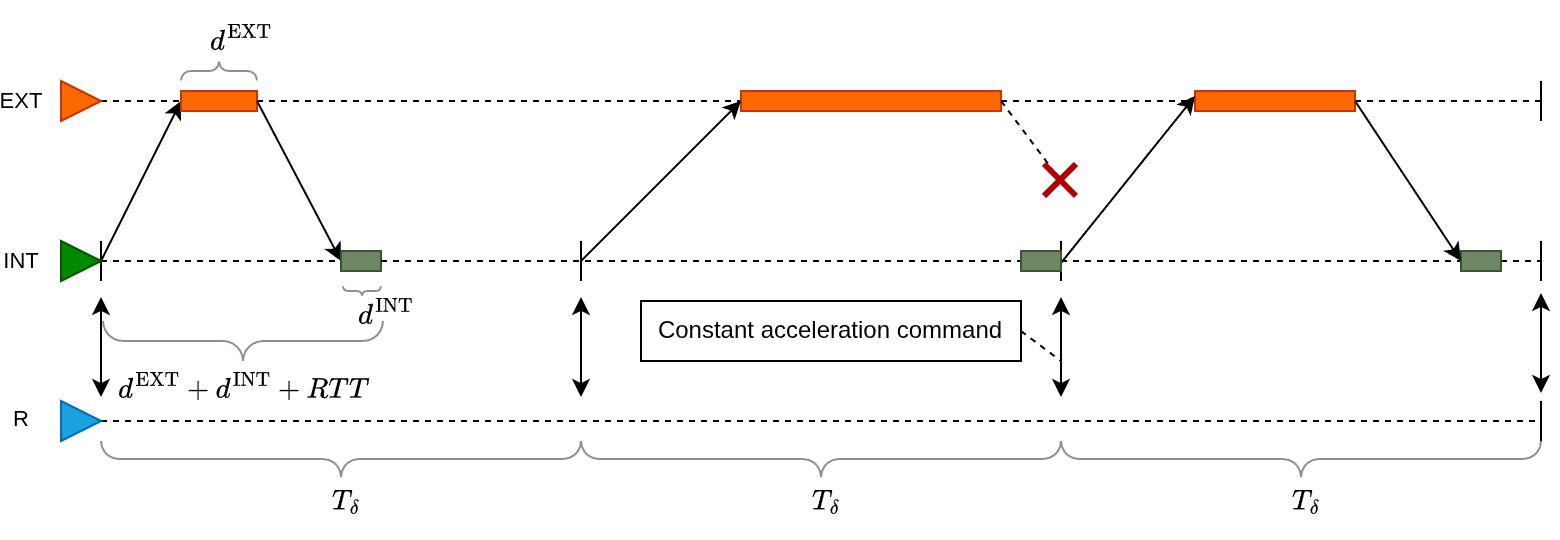}
  \caption{External control over the FCI (nominal synchronous model).
  Every $T_\delta$, INT measures the state of the robot (R) and applies the command
  returned by EXT one or more cycles earlier.
  The vertical arrows represent the periodic exchange of state and commands between INT and R (via an undisclosed interface); the diagonal arrows represent the transmission of states from INT to EXT and the corresponding command from EXT to INT through UDP.
  A command is applied only if its round trip meets the budget of \eqref{eq:budget}; otherwise it is discarded and INT holds the last commanded acceleration.}
  \label{fig:sequence}
\end{figure*}

The FCI connects the robot's onboard controller (INT in the following) to an external workstation (EXT) over UDP \cite{franka_robotics_franka_2025}.
Every $T_\delta \equiv \SI{1}{\milli\second}$, INT exchanges one packet pair with EXT, sending the current robot state and receiving the next command, and steps its internal controller \cite{franka_robotics_franka_2025}.
Two properties of this interface, illustrated in Fig.~\ref{fig:sequence}, are at the origin of the problem addressed in this work.
First, \texttt{libfranka} exposes \emph{blocking} reads only: the external process waits until a state packet arrives, so the EXT cycle is paced by INT.
Second, when operating in external control mode, EXT must supply one reference sample for \emph{every} INT cycle, since INT performs no onboard interpolation of sparse setpoints, unlike most industrial controllers.
For a command to be applied at a given cycle, its round trip must fit within the cycle: the documentation requires the sum of the network round-trip time (RTT), the execution time of the external loop ($d^{\mathrm{EXT}}$), and the INT's processing time ($d^{\mathrm{INT}}$) to stay below $T_\delta$,
\begin{equation}
  \mathrm{RTT} + d^{\mathrm{EXT}} + d^{\mathrm{INT}} \,<\, T_\delta,
  \label{eq:budget}
\end{equation}
and in practice recommends keeping $d^{\mathrm{EXT}} < \SI{0.5}{\milli\second}$ \cite{franka_robotics_franka_2025}.
Whenever the budget of \eqref{eq:budget} is exceeded for a cycle, FCI discards that command; INT then holds the last commanded acceleration constant and keeps running, and only after 20 consecutive discarded commands does it stop the session with a communication error \cite{franka_robotics_franka_2025}.
This extrapolation lets the actual and commanded references drift apart during loss bursts.

\subsection{Admissible References}
\label{sec:problem:fci-budget}

Upon receiving a reference, INT reconstructs the kinematic quantities it was not given and verifies them against the robot's limits.
Let $\Delta$ denote the backward finite-difference operator,
$$\Delta \mathbf x \triangleq \mathbf x(k)- \mathbf x (k-1),$$
with higher-order differences defined recursively as $\Delta^m\mathbf x(k)=\Delta\bigl(\Delta^{m-1}\mathbf x(k)\bigr)$.
For a joint position reference $\mathbf{q}_s(k)$, the case analyzed in this work, the reconstruction reduces to
\begin{equation}
  \begin{split}
    \dot{\mathbf q}(k) &= \frac{\Delta\mathbf q_s(k)}{T_\delta}, \\
    \ddot{\mathbf q}(k) &= \frac{\Delta\dot{\mathbf q}(k)}{T_\delta} = \frac{\Delta^2\mathbf q_s(k)}{T_\delta^2}, \\
    \dddot{\mathbf q}(k) &= \frac{\Delta\ddot{\mathbf q}(k)}{T_\delta} = \frac{\Delta^3\mathbf q_s(k)}{T_\delta^3},
  \end{split}
  \label{eq:backward_euler}
\end{equation}
from which the required actuation torques follow through the rigid-body dynamics \cite{featherstone_dynamics_2008}
\begin{equation}
  \begin{split}
    \boldsymbol\tau(k) &= \mathbf B\bigl(\mathbf q(k)\bigr)\,\ddot{\mathbf q}(k) \\
    &+ \mathbf C\bigl(\mathbf q(k),\dot{\mathbf q}(k)\bigr)\,\dot{\mathbf q}(k)
    + \mathbf f \bigl( \dot{\mathbf q}(k) \bigr)
    + \mathbf g\bigl (\mathbf q(k)\bigr),
  \end{split}
  \label{eq:dynamics}
\end{equation}
where $\mathbf B \in \mathbb R^{n \times n}$ is the inertia matrix, $\mathbf C \in \mathbb R^{n \times n}$ accounts for the centrifugal and Coriolis effects, $\mathbf f \in \mathbb R^n$ contains joint friction, $\mathbf g \in \mathbb R^n$ collects the gravitational torques, and $\mathbf q, \dot{\mathbf q}, \ddot{\mathbf q} \in \mathbb R^n$ respectively represent actual joint positions, velocities, and accelerations, with $n \in \mathbb N$ being the number of degrees of freedom\footnote{$n=7$ for the Panda robot.}.
INT verifies that positions, velocities, accelerations and jerks, as well as the torques in \eqref{eq:dynamics} and their rates, lie within their admissible ranges \cite{franka_robotics_franka_2025-1}; a violation aborts the motion with a discontinuity or limit error.
The admissible set of references is narrow.
Since \eqref{eq:backward_euler} divides successive differences by powers of $T_\delta$, a deviation $\epsilon$ of $\mathbf q_s(k)$ from the value that would preserve the previously established acceleration produces an additional jerk term of $\epsilon/T_\delta^3$.
At $T_\delta = \SI{1}{\milli\second}$, a deviation of a few microradians is enough to reach the jerk limit of the Panda \cite{franka_robotics_franka_2025-1}, and the other limits act analogously.
Informally, we experimentally confirmed this microradian-scale tolerance: perturbing an otherwise feasible trajectory with uniform noise on the order of $\SI[print-unity-mantissa=false]{1e-6}{\radian}$ was sufficient to trigger violations.

\subsection{The Synchronous ROS~2 Implementation}
\label{sec:problem:sync}

%

\begin{table}
    \centering
    \caption{Loop timings over the benchmark runs. 
    LF: bare \texttt{libfranka} motion-generation example; RS: official synchronous ROS~2 implementation executing a joint trajectory; RA: our asynchronous ROS~2 implementation executing the same joint trajectory as RS.
    Measurements are obtained with the LTTng tracer.}
    \label{tab:timings}
    \begin{tabular}{l|c|c|c}
         & \textbf{LF} & \textbf{RS} & \textbf{RA} \\
        \hline
        $d^{\mathrm{EXT}}$ median [\si{\milli\second}] & 0.012 & 0.065 & 0.013 \\
        $d^{\mathrm{EXT}}$ p99 [\si{\milli\second}]    & 0.040 & 0.137 & 0.048 \\
        $d^{\mathrm{EXT}}$ max [\si{\milli\second}]    & 0.087 & 2.174 & 0.141 \\
        unserved cycles [\si{\percent}]           & 0.18  & 5.91  & 0.41  \\
    \end{tabular}
\end{table}

\begin{figure}
    \centering
    \includegraphics[width=\linewidth]{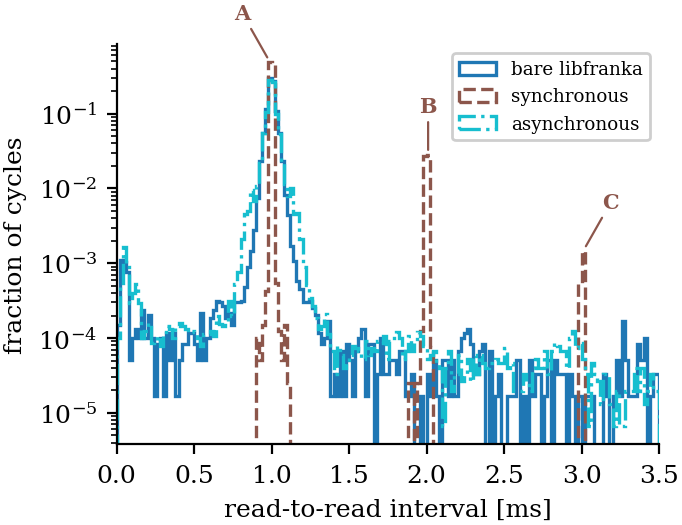}
    \caption{Distribution of the read-to-read interval, normalized by the number of measurements of each capture.
    For the synchronous implementation, the distribution peaks at multiples of $T_\delta$ (labeled as A, B and C).
    Measurements are obtained with the LTTng tracer.}
    \label{fig:timings:period}
\end{figure}
    
The official \texttt{franka\_ros2} hardware interface performs the blocking FCI state read inside the read phase of the \texttt{ros2\_control} main loop \cite{chitta_ros_control_2017}.
The latter is driven by its own periodic thread: at every cycle, it wakes on a workstation timer, performs the read phase, the controllers' updates and the write phase, then it sleeps until the next cycle boundary.
Because the FCI read is blocking, the read phase itself does not complete until a new state packet arrives from INT.
As a result, each iteration of the loop is paced by two different clocks: the beginning of the read phase is determined by the workstation timer (EXT), while its completion is determined by the arrival of the state packet generated by the robot controller (INT).
Since INT and EXT's clocks are independent, the relative phase between the two drifts slowly and leads to periodic unserved cycles.

Because the controllers execute between the read and the write, their computation lies on the critical path that \eqref{eq:budget} constrains.
The immediate cost of this arrangement is visible in Table~\ref{tab:timings}: the external ROS~2 control loop (RS) takes a median of \SI{0.065}{\milli\second}, against \SI{0.012}{\milli\second} for a bare \texttt{libfranka} program (LF) driving the same robot, and reaches \SI{2.174}{\milli\second} in the worst cycle, exceeding $T_\delta$ and leading to the discard of the command of that cycle.
It is worth highlighting that the controllers exercised for these measurements only forward joint position commands; critically, the margin is expected to shrink as the application becomes more demanding.

A more damaging consequence is produced by the drifting phase between the two clocks.
When the EXT timer wakes shortly before receiving the next state packet from INT, the blocking read returns after a short wait and the cycle completes comfortably, producing a read-to-read interval close to $T_\delta$ (Fig.~\ref{fig:timings:period}, peak A).
Conversely, when the EXT timer wakes immediately after a state packet has already been sent by INT, the read must wait almost a full period $T_\delta$ for the following packet, since it returns only on a fresh state.
This causes the ROS~2 cycle to overrun, and the next completed read will happen on a future wake of the EXT timer (as in Fig.~\ref{fig:timings:period}, peaks B and C).
When the EXT cycle overruns, one or more INT cycles remain unserved.
During those cycles, INT holds the last commanded acceleration, as described in Sec.~\ref{sec:problem:fci}, thus distorting the intended reference.
In our benchmarks (Table~\ref{tab:timings}), RS results in \SI{5.91}{\percent} unserved cycles, against \SI{0.18}{\percent} for LF.


\subsection{Reference Distortion under Sampling Jitter}
\label{sec:problem:jitter}

\begin{figure}
  \centering
  \includegraphics[width=\linewidth]{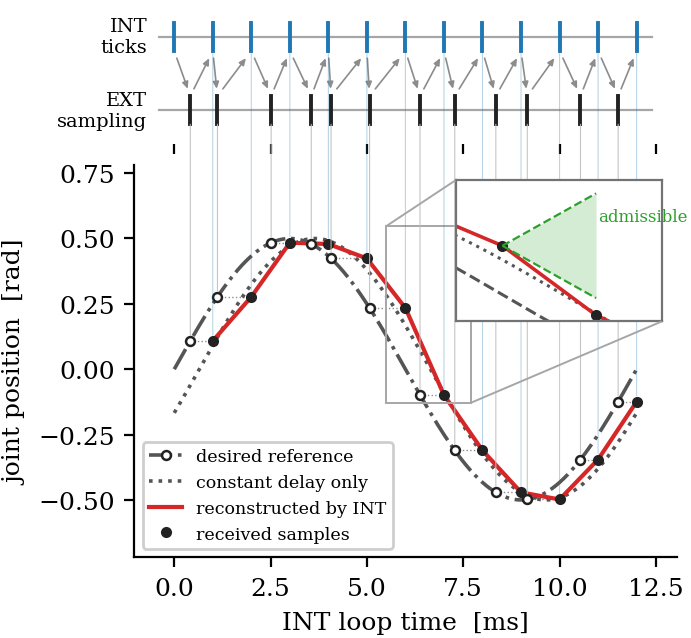}
  \caption{Reference distortion under sampling jitter.
  EXT samples the trajectory after each INT tick, at instants whose
  jitter displaces the samples (dotted lines); INT files them on its
  uniform grid and differences them, so the reconstructed reference
  (red) departs from the desired one (dashed).
  On a steep segment (inset) the resulting increment leaves the
  admissible cone and aborts the motion.
  Jitter amplitude exaggerated for legibility.}
  \label{fig:jitter}
\end{figure}

Careful real-time configuration can reduce the number of unserved cycles, but a second effect survives it: on the cycles that are served, the reference itself is distorted by \emph{sampling jitter}.
In a typical ROS~2 deployment, a planned trajectory is sampled using the EXT workstation clock, which is typically a soft real-time machine communicating with INT through standard TCP/IP protocols.
Network round trip, the non-deterministic latency of the operating-system network stack and scheduler on EXT, and the processing time of INT are all irreducible causes of sampling jitter \cite{reghenzani_real-time_2019,casini_response-time_2019}.

More formally, let $\mathbf{q}^*(t)$ be the planned trajectory, intended as a continuous function; the reference applied by INT at cycle $k$ is the trajectory evaluated by EXT at some instant $t_s(k)$, i.e., $\mathbf{q}^*\bigl( t_s(k) \bigr)$, but \eqref{eq:backward_euler} treats it as the uniform sample taken at $k T_\delta$.
Even if the two clocks were perfectly aligned and INT were perfectly real-time, the jitter sources listed above would induce a time-domain sampling error of $\delta t(k)=t_s(k)-k T_\delta$.
By first order Taylor expansion $\mathbf{q}_s(k)\approx\mathbf{q}^*(kT_\delta)+\dot{\mathbf{q}}^*(kT_\delta)\delta t(k)$, therefore the sampling jitter induces an error in the finite differences calculated by INT.
For the velocity,
\begin{equation}\label{eq:velocity_error}
    \dot{\mathbf q}(k)
      = \frac{\Delta\mathbf q_s(k)}{T_\delta} 
      \simeq \underbrace{%
          \frac{\Delta \mathbf q^*(k T_\delta)}{T_\delta}
        }_{\text{intended velocity}} 
      + \underbrace{\dot{\mathbf q}^*(k T_\delta)
        \frac{\Delta \delta t(k)}{T_\delta}}_{%
          \text{sampling error}}.
\end{equation}
By chaining the finite differences of \eqref{eq:backward_euler}, the error induced by sampling jitter on the calculated jerk is $\dot{\mathbf q}^*\Delta^3\delta t/T_{\delta}^3$.
At $\dot q= \SI[per-mode=symbol]{1}{\radian\per\second}$, a single sample displaced by \SI{4}{\micro\second} displaces the commanded position by \SI{4}{\micro\radian} and produces a jerk error of \SI[per-mode=symbol]{1.2e4}{\radian\per\second\cubed}, above the jerk limit of every joint of the robot \cite{franka_robotics_franka_2025-1}.
In the synchronous implementation, where the sample is taken as soon as the packet arrives, the problem is amplified as $t_s(k)$ inherits the arrival jitter directly.
This is shown in Fig.~\ref{fig:jitter} and it explains why, in our tests, discontinuity aborts coincided with the cycles carrying the longest inter-arrival gaps.

Sampling the trajectory with INT's own timestamps, i.e., the robot time carried by each state packet, removes this effect, and in our tests eliminated all violations; it requires, however, the trajectory in closed form and an external loop running exactly at INT's rate, neither of which a controller-agnostic ROS~2 stack running on a general-purpose OS can assume.

To work around the problem, \texttt{libfranka} offers a rate limiter and a low-pass filter, but neither addresses its cause: the rate limiter only saturates the command derivatives to the admissible ranges, trading the discontinuity aborts for sustained link oscillations, while the low-pass filter attenuates the distortion without removing it, leaving audible motor noise and sporadic aborts.

\subsection{Robustness of the Velocity and Torque Interfaces}
\label{sec:problem:masking}

The same jitter affects velocity and torque commands, yet these interfaces appear to work reliably, which explains why the problem has drawn little attention and why existing frameworks simply advise against the position interface \cite{skerlj_bridging_2026}.
The reason follows from \eqref{eq:backward_euler}.
Repeating the calculation of Sec.~\ref{sec:problem:jitter} for the velocity command interface, one obtains that the error induced on the calculated jerk is $\ddot{\mathbf{q}}^*\, \Delta^2\delta t /T_\delta^2$: the sensitivity to the same timing error is reduced, making the position interface the most sensitive one.
Another reason this problem is often unnoticed is that, when using the velocity or torque commands to control position, the loop closure itself absorbs the residual distortion as a disturbance.
The scheme of Sec.~\ref{sec:architecture:refgen} recovers both mechanisms within the position interface: the motion is driven by the commanded velocity, so that the increment reaching INT is not the difference of two displaced position samples, and the position command acts only through a loop closed on the robot's own commanded state.


\section{PROPOSED ARCHITECTURE}
\label{sec:architecture}

\begin{figure*}
    \centering
    \includegraphics[width=\linewidth]{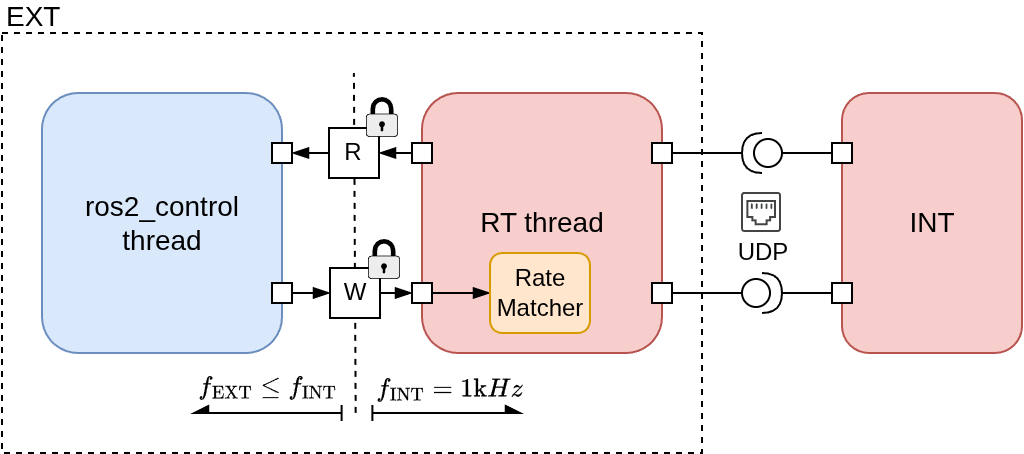}
    \caption{Proposed architecture.
    The \texttt{ros2\_control} thread and the real-time thread run on the workstation (EXT); they exchange state and command through mutex-protected buffers, so ROS-side jitter never reaches the robot.
    The real-time thread talks to the onboard controller (INT) over the UDP link, and matches a slower command stream to \SI{1}{\kilo\hertz} through the rate-matching stage on the write path.}
    \label{fig:architecture}
\end{figure*}

The proposed stack, summarized in Fig.~\ref{fig:architecture}, rests on two open-source components: a community-maintained \texttt{libfranka} repository \cite{tingelstad_libfranka_2024} that brings the modern control API to the Panda, extended with the Panda kinematic limits \cite{franka_robotics_franka_2025-1}, and a ROS~2 Jazzy package built on top of it providing a hardware interface for the robot \cite{franka_robotics_franka_ros2_2025}.
On this basis, Sec.~\ref{sec:architecture:async} introduces the asynchronous communication thread that decouples the \texttt{ros2\_control} loop from the FCI channel; Sec.~\ref{sec:architecture:filtering} exploits this decoupling to run the \texttt{ros2\_control} loop at a lower rate, matched to the robot loop by a rate matching stage applicable to any command interface; and Sec.~\ref{sec:architecture:refgen} presents the position-domain reference generation that makes the position interface reliable.

\subsection{Asynchronous Communication Thread}
\label{sec:architecture:async}

The asynchronous interface addresses the problem of Sec.~\ref{sec:problem:sync} by confining FCI communication to a dedicated real-time thread, scheduled under \texttt{SCHED\_FIFO} at the highest priority (99) and only containing the communication pipeline.
The thread runs its own loop paced by the robot: each iteration blocks on the state read, publishes the received state into a mutex-protected buffer, fetches the next command from a second mutex-protected buffer, and transmits it within the budget of \eqref{eq:budget}.
On the \texttt{ros2\_control} side, the logic reduces to buffer exchanges: the read phase copies the latest state and the write phase deposits the latest command.
ROS-side jitter --- being due to scheduling, DDS activity or controller cost --- therefore terminates at the buffers and never reaches the FCI channel; conversely, a late ROS cycle no longer implies a missed robot cycle, since the thread keeps serving INT with a consistent command.
The measurements of Table~\ref{tab:timings} quantify both effects: thanks to the dedicated thread, $d^{\text{EXT}}$ has a median of \SI{0.013}{\milli\second}, comparable to the \SI{0.012}{\milli\second} of a bare \texttt{libfranka} program; the synchronous interface produces \SI{5.91}{\percent} unserved cycles while our asynchronous implementation only produces \SI{0.41}{\percent}.  
Furthermore, with the controllers moved off the real-time path, the delay added to the FCI exchange does not scale with the complexity of the application.
In Fig.~\ref{fig:timings:period} the quantization of the synchronous interface gives way to the broad, continuous distribution of the bare \texttt{libfranka} loop: the thread is paced by the arrival of the packets rather than by a workstation timer, and the phase competition of Sec.~\ref{sec:problem:sync} does not arise.
 
Robustness to packet loss is handled at the same level.
The thread timestamps every state read and compares the measured inter-arrival period $\hat T(k)$ against $T_\delta$; whenever $|\hat T(k) - T_\delta| > \SI{0.25}{\milli\second}$, the cycle is flagged as anomalous, the write is skipped, and the last consistent command is held, preventing a command computed against a stale or duplicated state from being differentiated by \eqref{eq:backward_euler}.
Position-type modes are additionally gated at activation: no command is transmitted until both the thread and the ROS-side loop have consumed a fresh state, avoiding first-cycle steps at controller switches.

Because the timing requirement is confined to the communication thread, and the processing in the latter is reduced to a bare minimum, the stack does not enforce the use of a real-time kernel patch: the experiments of Sec.~\ref{sec:validation} are obtained on a \texttt{PREEMPT\_RT}-patched kernel, while the applications in Sec.~\ref{sec:applications} run with \texttt{SCHED\_FIFO} scheduling alone in Salerno, and on a RT-patched one at IDSIA.

\subsection{Rate Matching for Slower Control Loops}
\label{sec:architecture:filtering}

The proposed architecture admits control loops slower than what FCI prescribes: the \texttt{ros2\_control} loop may run at any rate $f_\text{EXT} \le f_\text{INT}$, which relaxes its real-time requirements and matches the native rate of most command sources, from motion planners to teleoperation devices.
The rate-matching stage described here is what makes this possible.

A source running at $f_\text{EXT}$ delivers one update every $N = \lceil f_\text{INT}/f_\text{EXT} \rceil$ robot cycles.
Passing those command updates on as they arrive would hold each value for $N-1$ cycles and step on the $N$-th, and \eqref{eq:backward_euler} would turn that staircase into impulsive motor actions.
To prevent this, a rate-matching moving-average filter is run on the communication thread to smooth the reference without distorting it.
By choosing the window of the filter as $N = \lceil f_\text{INT}/f_\text{EXT} \rceil$, each command update is distributed across the $N$ cycles separating it from the next, and the staircase becomes a piecewise-linear ramp.
The filter introduces a group delay of $\frac{N-1}{2}T_\delta=(N-1)\times$\SI{0.5}{\milli\second}.
The delay is limited to a few milliseconds assuming a reasonable value of $f_{\text{EXT}}\ge\SI{100}{\hertz}$.
Because it acts on the command samples irrespective of their meaning, the filter applies to any command interface.

\subsection{Position-Domain Reference Generation}
\label{sec:architecture:refgen}

The analysis of Sec.~\ref{sec:problem:jitter} indicates that any scheme sampling the trajectory with the workstation clock inherits sampling jitter and, over time, clock drift.
As a consequence, the proposed control scheme abandons external time references and recovers the two mechanisms that make the velocity and torque interfaces robust (Sec.~\ref{sec:problem:masking}).

Let $\mathbf q_d(k)$ be the last commanded position reported by INT in the state packet; $\mathbf q_{\mathrm{ext}}(k)$ be the position command (raw or filtered) most recently deposited by the ROS side and $\dot{\mathbf q}_{\mathrm{ext}}(k)$ be its feed-forward velocity obtained by finite-differencing successive position commands.
The proposed method generates the reference in the \emph{position domain} from the latest available position command
\begin{equation}
  \mathbf q_s(k) \,=\, \mathbf q_d(k)
  + T_\delta\,\dot{\mathbf q}_{\mathrm{ext}}(k)
  + \lambda\bigl(\mathbf q_{\mathrm{ext}}(k) - \mathbf q_d(k)\bigr),
  \label{eq:refgen}
\end{equation}
where $\lambda$ is a tunable parameter.

Equation~\eqref{eq:refgen} has three properties that address the failure modes of Sec.~\ref{sec:problem}: (i) the motion is driven by the feed-forward term $T_\delta\dot{\mathbf q}_{\mathrm{ext}}(k)$ while the position enters only through a proportional term closing the loop on $\mathbf q_d(k)$, so the position inherits the sensitivity of the velocity interface; (ii) if the external stream stalls, the reference degrades gracefully toward the last target instead of stepping; (iii) no timestamp of either machine appears in it, making it immune to clock drift.

An analogous scheme covers the Cartesian pose interface: the commanded pose is propagated by the externally commanded twist over one period, and a proportional twist correction toward the externally commanded pose is applied incrementally.

Since the scheme already delivers a continuous command stream, the command low-pass filter offered by \texttt{libfranka} is redundant, and it must in fact be disabled: applied to an incremental command it scales the increment itself, and with it both the feed-forward term and the correction gain, so that the robot follows a fraction of the commanded motion (Sec.~\ref{sec:validation}).

We do not claim formal guarantees for the proposed scheme; its behavior is assessed empirically in our experiments.


\section{EXPERIMENTAL VALIDATION}
\label{sec:validation}

\subsection{Setup}
\label{sec:validation:setup}
 
The experiments reported in this section are performed on a Panda controlled from a workstation running Ubuntu~24.04 (patched with \texttt{PREEMPT\_RT} \cite{reghenzani_real-time_2019}) and ROS~2 Jazzy, connected to the robot by a dedicated Ethernet link.
We compare the official (synchronous) \texttt{franka\_ros2} hardware interface ported to ROS~2 Jazzy with our asynchronous interface.
 
Joint trajectories are produced by a robot-agnostic reference generator \cite{risi_simplifying_2026}, which validates and interpolates the waypoints and publishes the resulting command stream to the controller under test.
The same $\sim \SI{18.32}{\second}$ multi-joint trajectory is replayed in every run.
 
Loop timings are collected with LTTng user-space tracepoints placed around the state read and the command write, as reported in Sec.~\ref{sec:problem:sync} and Sec.~\ref{sec:architecture:async}; robot states and the trajectory results of each run are recorded from the ROS~2 graph.
 
\subsection{Position Control}
\label{sec:validation:tracking}

We evaluate the performance of the proposed position reference generation scheme in \eqref{eq:refgen} by comparing executions of the same trajectory with both the synchronous and asynchronous interfaces.
For the synchronous interface, the low-pass filter is active as it is required to safely control the arm via position commands. 
For the asynchronous interface, we compare the behavior with and without the low-pass filter and set the control frequency to $f_{\text{EXT}}=f_{\text{INT}}=\SI{1}{\kilo\hertz}$.
All three configurations have the rate limiter enabled to prevent communication errors.
For the experiments we set $\lambda = \num{1e-3}$.

\begin{figure}
  \centering
  \includegraphics[width=\linewidth]{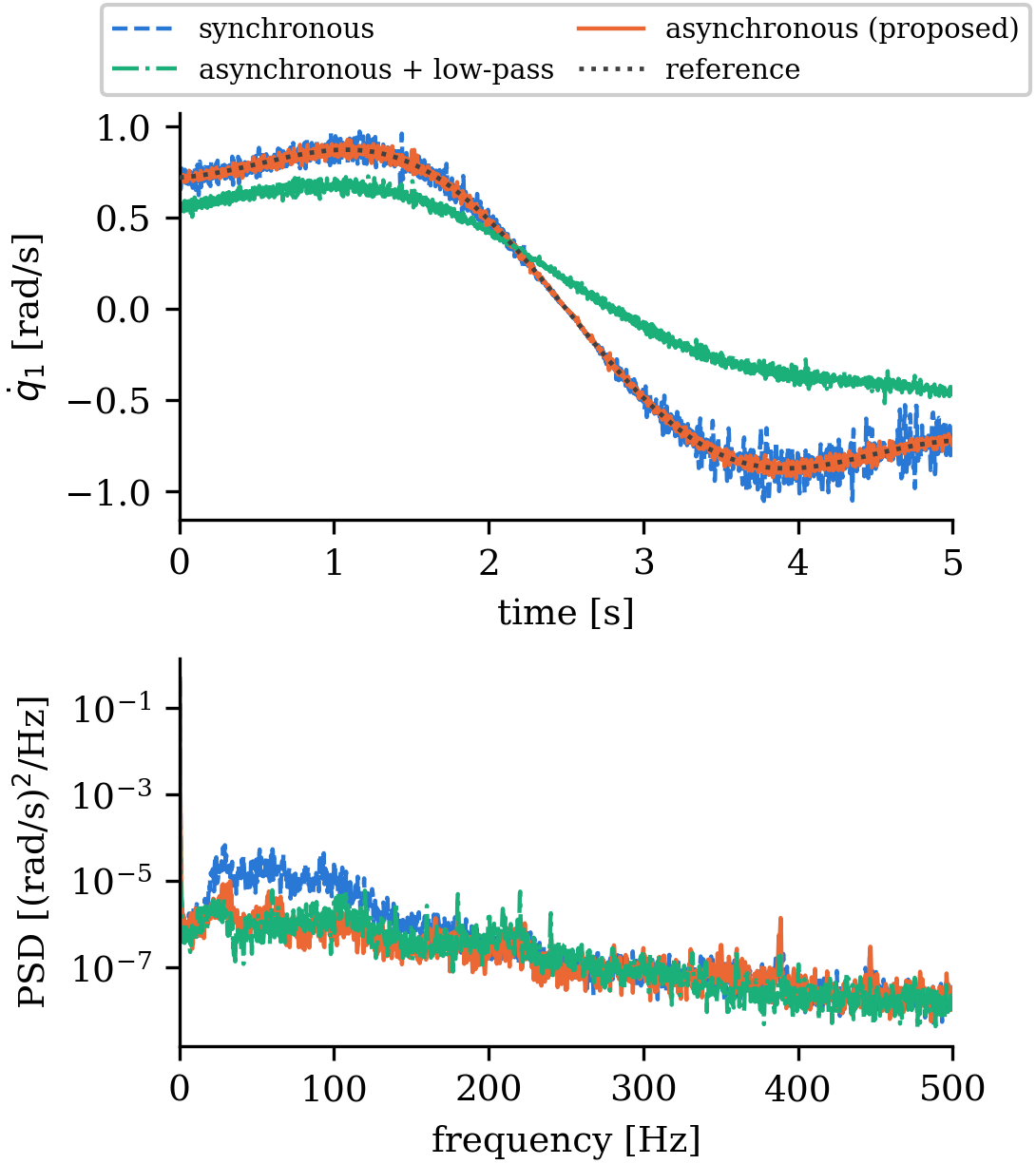}
  \caption{Measured joint velocity (top) and its power spectral density
  (bottom) on joint 1, for the synchronous configuration,
  the proposed asynchronous one, and the proposed one with the low-pass filter
  enabled.
  The synchronous velocity tracks the reference, but is visibly rough; the
  filtered one is smooth, but follows a fraction of the commanded motion; the
  proposed one tracks and stays smooth.}
  \label{fig:tracking}
\end{figure}

Fig.~\ref{fig:tracking} compares the measured joint velocity and its spectrum for three configurations: synchronous, the proposed asynchronous configuration, and the proposed one with the addition of the low-pass filter.
The synchronous interface follows the reference but carries an order of magnitude more spectral power than the other two configurations in the $20-150$\,Hz band: this is the roughness the rate limiter leaves behind once it has removed the invalid commands, and it manifests as audible motor noise during the execution of the trajectory.
The low-pass filter suppresses that excess but, as discussed in Sec.~\ref{sec:architecture:refgen}, it hinders the tracking by scaling the incremental command (visible as the shrunken joint velocity curve in Fig.~\ref{fig:tracking}).
The proposed configuration attains faithful tracking while keeping the low band power of the filtered case.
The remaining joints behave consistently\footnote{Results for all joints are available at \url{https://sites.google.com/unisa.it/fer-ros2/results}}.

Each configuration is evaluated over 10 repetitions of the same trajectory.
The oscillatory behavior of the synchronous interface is accompanied by frequent robot-side protective stops (due to exceeding velocity/acceleration discontinuities or effort/power limits), with only 2/10 runs completing successfully, whereas the best asynchronous configuration showcases 10/10 completion rate.

\subsection{Rate Matching}
\label{sec:validation:ratematching}
 
To validate the rate matching mechanism described in Sec.~\ref{sec:architecture:filtering}, the robot is driven from a velocity command source running at $f_{\text{EXT}} = \SI{125}{\hertz}$ (hence $N = 8$) and we measure what reaches the robot with the rate-matching stage enabled and disabled.

\begin{figure}
  \centering
  \includegraphics[width=\linewidth]{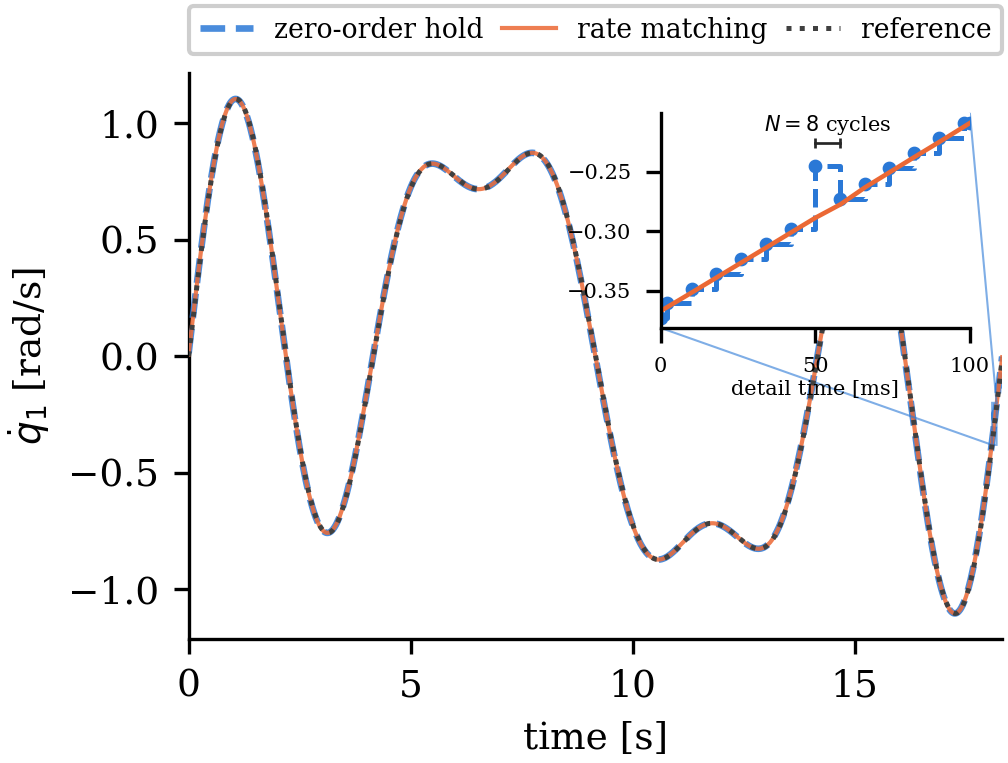}
  \caption{Commanded joint velocity reaching the robot from a \SI{125}{\hertz} source,
  with rate matching disabled and enabled. 
  The two commands, from separate runs, are aligned to the reference.
  The envelopes coincide over the trajectory (main), while the detail shows the
  zero-order-held staircase, whose single riser spans the $N=8$ cycles that the
  rate-matched command crosses as a ramp.}
  \label{fig:ratematching}
\end{figure}

Results are shown in Fig.~\ref{fig:ratematching}: with the stage disabled, the command is the zero-order hold of the source, constant for $N-1$ cycles and stepping on the $N$-th; the single step carries the whole update, so the per-cycle increment presented to INT is $N$ times the intended one.
With the stage enabled, the same update is spread across the $N$ cycles it spans and the command becomes a ramp whose per-cycle increment is the intended one, at the analytical cost of a group delay of $\frac{N-1}{2}T_\delta = \SI{3.5}{\milli\second}$.
 
\section{APPLICATIONS}
\label{sec:applications}

We demonstrate the proposed stack on four applications, each exercising a capability that collaborative robotics commonly relies on:

\begin{enumerate}
\item collision-aware motion planning and execution via \texttt{MoveIt!}\ and ROS2-native controllers (Sec.~\ref{sec:experiment-moveit});\label{itm:experiment-moveit}
\item compliance control for physical human-robot interaction, at standstill and along trajectories, with and without human perturbation (Sec.~\ref{sec:experiment-compliance-control});\label{itm:experiment-compliance-control}
\item an end-to-end, \emph{position}-controlled pick-and-place (Sec.~\ref{sec:experiment-pick-and-place});\label{itm:experiment-pick-and-place}
\item haptic teleoperation of a mobile manipulator (Sec.~\ref{sec:experiment-teleoperation}).\label{itm:experiment-teleoperation}
\end{enumerate}

Together they exercise the position, velocity and torque command interfaces of our hardware interface.
The \texttt{MoveIt!}\ planning and compliance control applications are run on two different Panda setups, hosted in Salerno and at IDSIA, while pick-and-place and teleoperation are run in Salerno only, showing that the same software runs unchanged across different setups and machines.
Moreover, to ease the shake-out of the system for new users, we have restored all the basic \texttt{libfranka} examples\footnote{\url{https://github.com/unisa-acg/libfranka}} which, for the sake of brevity, are shown on our website\footnote{\url{https://sites.google.com/unisa.it/fer-ros2/libfranka}}.
It is worth highlighting that, in Salerno, the applications are run with a non-realtime kernel, to demonstrate that, as mentioned in Sec.~\ref{sec:architecture:async}, patching the system with \texttt{PREEMPT\_RT} is not a strict requirement as it was for the official release of \texttt{libfranka}.

\subsection{MoveIt!\ Planning and Execution}
\label{sec:experiment-moveit}

\begin{figure}
    \centering
    \begin{subfigure}[b]{0.32\columnwidth}
        \centering
        \includegraphics[width=\linewidth]{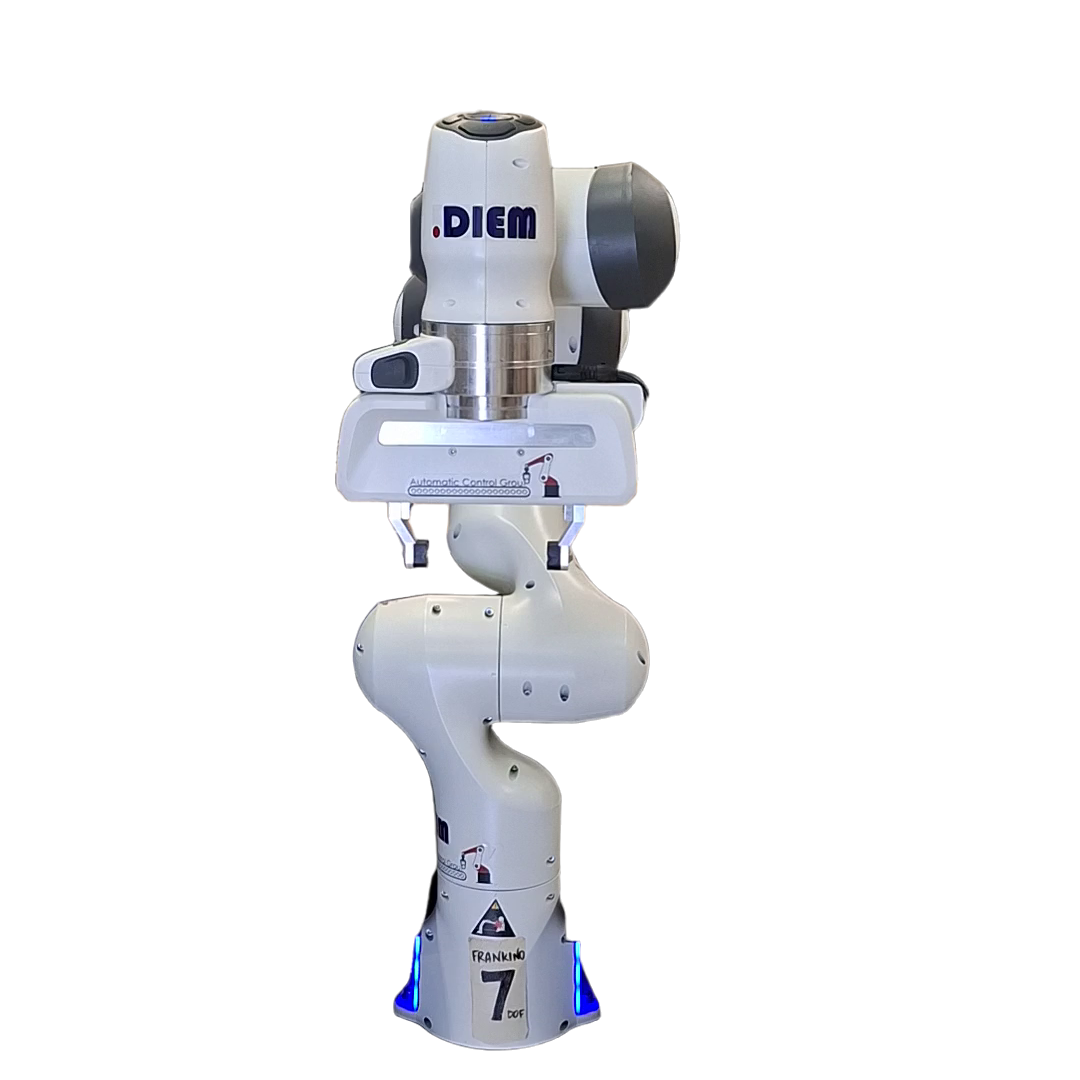}
    \end{subfigure}
    \hfill
    \begin{subfigure}[b]{0.32\columnwidth}
        \centering
        \includegraphics[width=\linewidth]{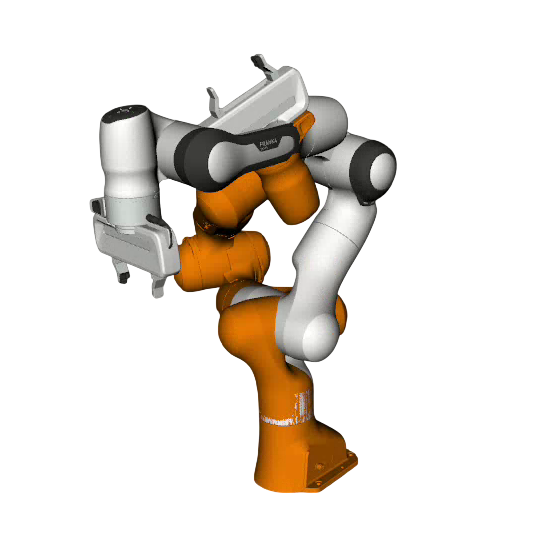}
    \end{subfigure}
    \hfill
    \begin{subfigure}[b]{0.32\columnwidth}
        \centering
        \includegraphics[width=\linewidth]{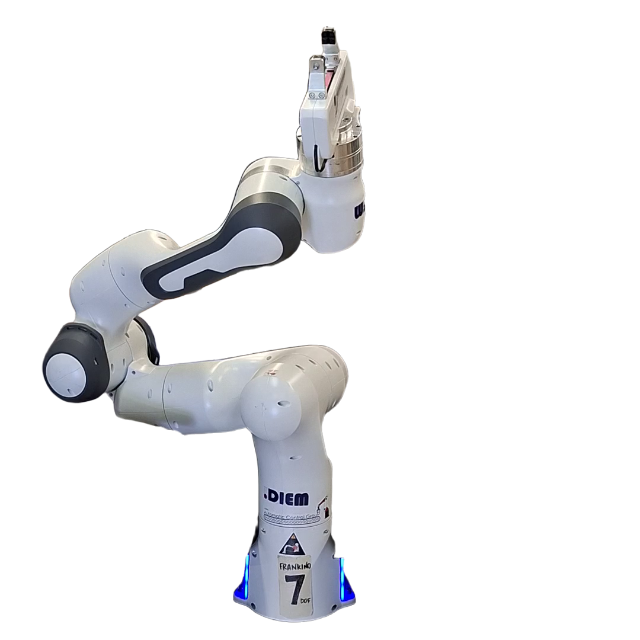}
    \end{subfigure}
        \\[1mm]
    \begin{subfigure}[b]{0.32\columnwidth}
        \centering
        \includegraphics[width=\linewidth]{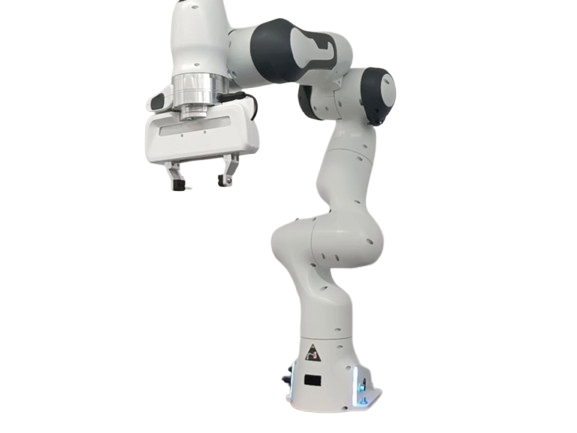}
        \caption{Initial pose}
        \label{fig:moveit-initial-pose}
    \end{subfigure}
    \hfill
    \begin{subfigure}[b]{0.32\columnwidth}
        \centering
        \includegraphics[width=\linewidth]{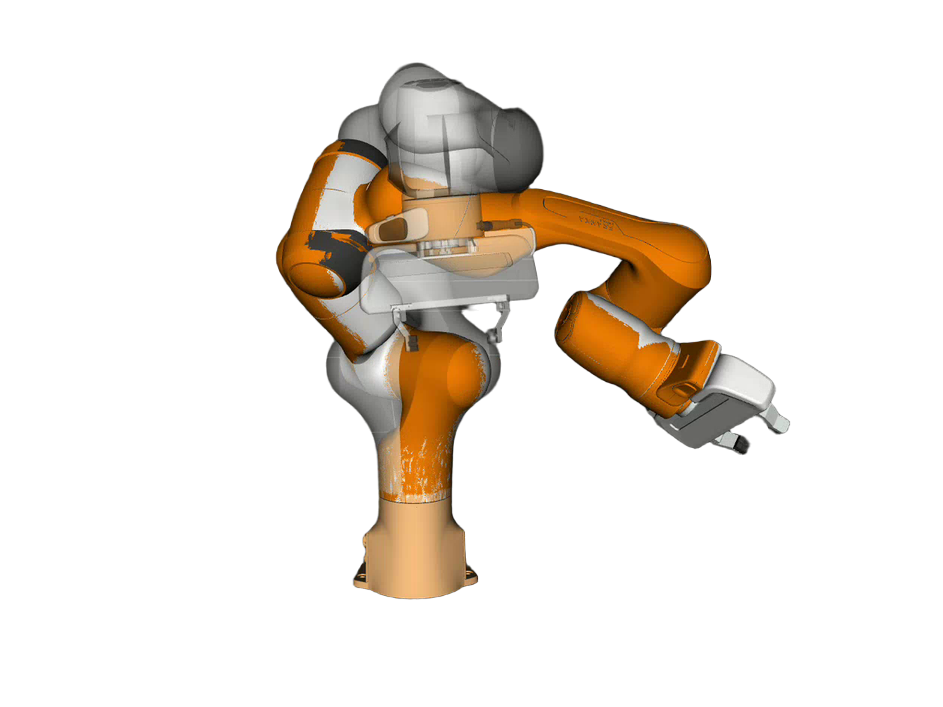}
        \caption{\texttt{MoveIt!} plan}
        \label{fig:moveit-plan}
    \end{subfigure}
    \hfill
    \begin{subfigure}[b]{0.32\columnwidth}
        \centering
        \includegraphics[width=\linewidth]{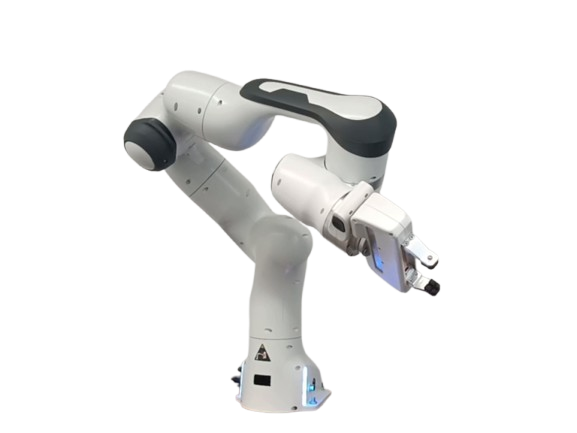}
        \caption{Final pose}
        \label{fig:moveit-final-pose}
    \end{subfigure}
    \caption{\texttt{MoveIt!} planning and execution, reproduced on both Panda setups: (top) Salerno; (bottom) IDSIA. Full videos at \url{https://sites.google.com/unisa.it/fer-ros2applications/moveit}}
    \label{fig:moveit}
\end{figure}

\begin{table}
\centering
\caption{Control parameters: the \textbf{Sec.} column points to the subsection presenting the application; in the \textbf{Axis} column, numbers indicate joints, while $\{x,y,z\}$ and $\{R,P,Y\}$ represent Cartesian-space linear and angular components respectively.}
\label{tab:control-parameters}
\setlength{\tabcolsep}{2.5pt}
\footnotesize
\begin{tabular}{c|c|c|c|c}
\textbf{Sec.} & \textbf{Axis} & $\mathbf K_P$ & $\mathbf K_I$ & $\mathbf K_D$ \\
\hline
\multirow{4}{*}{\ref{sec:experiment-moveit}} & 1,2,3,4 & \num{600} & \num{0} & \num{30} \\
 & 5 & \num{250} & \num{0} & \num{10} \\
 & 6 & \num{150} & \num{0} & \num{10} \\
 & 7 & \num{50} & \num{0} & \num{5} \\
\hline
\multirow{2}{*}{\ref{sec:experiment-compliance-control}\,\ref{itm:experiment-compliance-controll-standstill-stiff},\ref{itm:experiment-compliance-controll-trajectory-stiff}} & $x,y,z$ & \num{1000} & -- & \num{100} \\
 & $R,P,Y$ & \num{100} & -- & \num{10} \\
\hline
\multirow{2}{*}{\ref{sec:experiment-compliance-control}\,\ref{itm:experiment-compliance-controll-standstill-compliant},\ref{itm:experiment-compliance-controll-trajectory-compliant}} & $x,y,z$ & \num{100} & -- & \num{10} \\
 & $R,P,Y$ & \num{10} & -- & \num{5} \\
\hline
\multirow{2}{*}{\ref{sec:experiment-pick-and-place}} & $x,y,z$ & \num{10} & -- & -- \\
 & $R,P,Y$ & \num{10} & -- & -- \\
\hline
\multirow{3}{*}{\ref{sec:experiment-teleoperation}} & $x,y,z$ & \num{900} & -- & \num{100} \\
 & $R$ & \num{90} & -- & \num{10} \\
 & $P,Y$ & \num{90} & -- & \num{1}
\end{tabular}
\end{table}

This application exercises the \texttt{MoveIt!}\ planning and execution pipeline (Fig.~\ref{fig:moveit}).
Starting from the initial configuration (Fig.~\ref{fig:moveit-initial-pose}), the robot plans a motion to a randomly sampled goal pose (Fig.~\ref{fig:moveit-plan}) and executes it (Fig.~\ref{fig:moveit-final-pose}) through an effort-based \texttt{joint\_trajectory\_controller} tuned with the PID gains of Table~\ref{tab:control-parameters}.
This confirms that the stack effortlessly integrates with the standard ROS~2 planning tools, letting users move the robot to any feasible desired pose defined by the target application.

\subsection{Compliance Control}
\label{sec:experiment-compliance-control}

This application deploys a custom torque controller, named the \textit{compliance controller}, which applies a Cartesian-space PD control law at the end-effector, whose gains set how compliant the arm is to human interaction:
\begin{equation}\label{eq:compliance-control}
\begin{split}
    \boldsymbol\tau &= \mathbf J^\top(\mathbf q) ( \mathbf K_P \tilde{\mathbf x} + \mathbf K_D \dot{\tilde{\mathbf x}}) + \mathbf C(\mathbf q, \dot{\mathbf q}) \dot{\mathbf q} + \mathbf f(\dot{\mathbf q}) + \mathbf g(\mathbf q) \\
    &+ \bigl( \mathbf I_n - \mathbf J^\top(\mathbf q) \mathbf J^\dagger(\mathbf q) \bigr) \bigl( k_0 (\mathbf q_0 - \mathbf q) - 2\sqrt{k_0} \dot{\mathbf q}\bigr)
\end{split}
\end{equation}
where $\mathbf x$ and $\dot{\mathbf x}$ are the end-effector pose and twist, $\tilde\bullet \triangleq \bullet_d - \bullet$ indicates the error, and $\bullet_d$ represents the desired quantity.
$\mathbf I_n$ is the $n \times n$ identity matrix, and $\mathbf J(\mathbf q) \in \mathbb R^{6 \times n}$ is the configuration-dependent Jacobian matrix, with $\dagger$ indicating the pseudo-inverse of its transpose, while $k_0 \coloneqq 1.0$ and $\mathbf q_0 \coloneqq [0, 0, 0, \SI{-90}{\degree}, 0, \SI{107}{\degree}, 0]^\top$ are the nullspace stiffness and setpoint, respectively.

We exercise this controller in four different setups, with control gains $\mathbf K_P$ and $\mathbf K_D$ set according to Table~\ref{tab:control-parameters}:
\begin{enumerate}[label=\Alph*),ref=\Alph*]
    \item standstill human perturbation with stiff behavior;\label{itm:experiment-compliance-controll-standstill-stiff}
    \item standstill human perturbation with compliant behavior;\label{itm:experiment-compliance-controll-standstill-compliant}
    \item trajectory tracking with stiff behavior;\label{itm:experiment-compliance-controll-trajectory-stiff}
    \item trajectory tracking with compliant behavior and human perturbation.\label{itm:experiment-compliance-controll-trajectory-compliant}
\end{enumerate}

\begin{figure*}
    \centering
    \begin{subfigure}[b]{0.48\textwidth}
        \centering
        \includegraphics[width=0.49\linewidth]{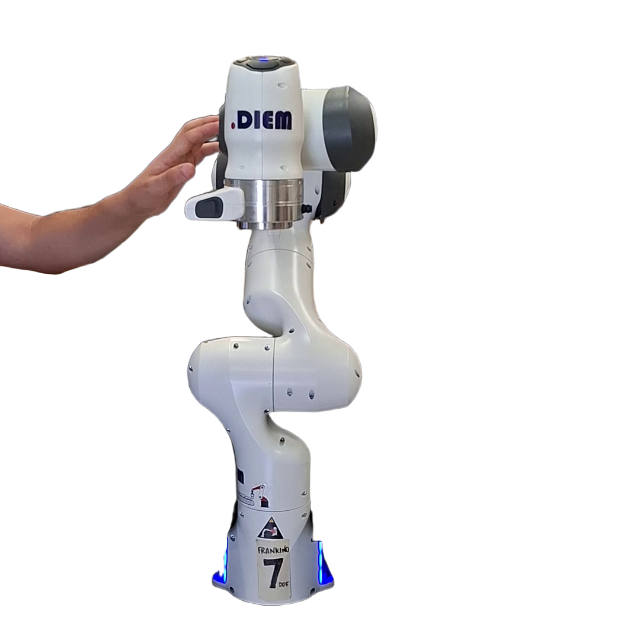}\hfill
        \includegraphics[width=0.49\linewidth]{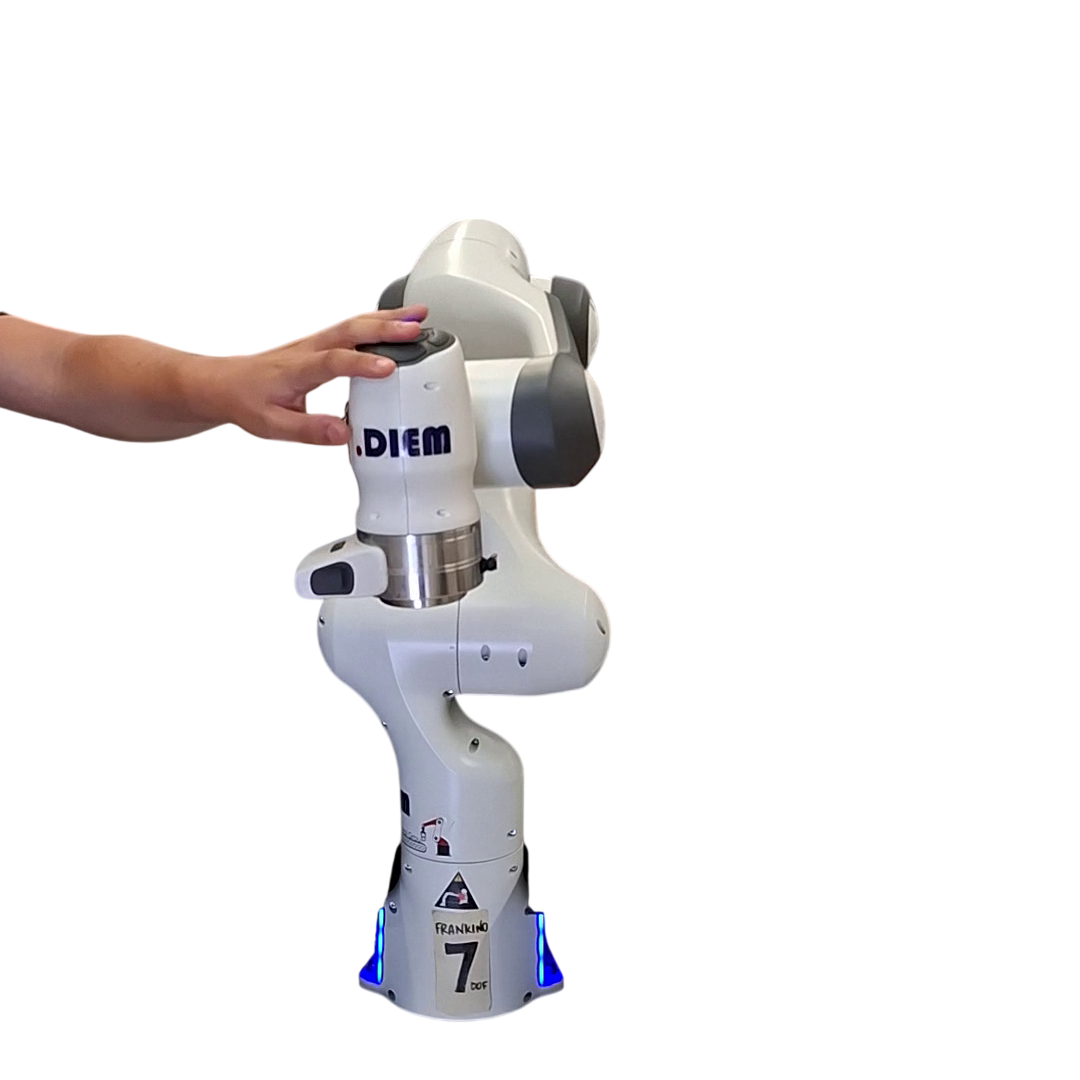}\\[1mm]
        \includegraphics[width=0.49\linewidth]{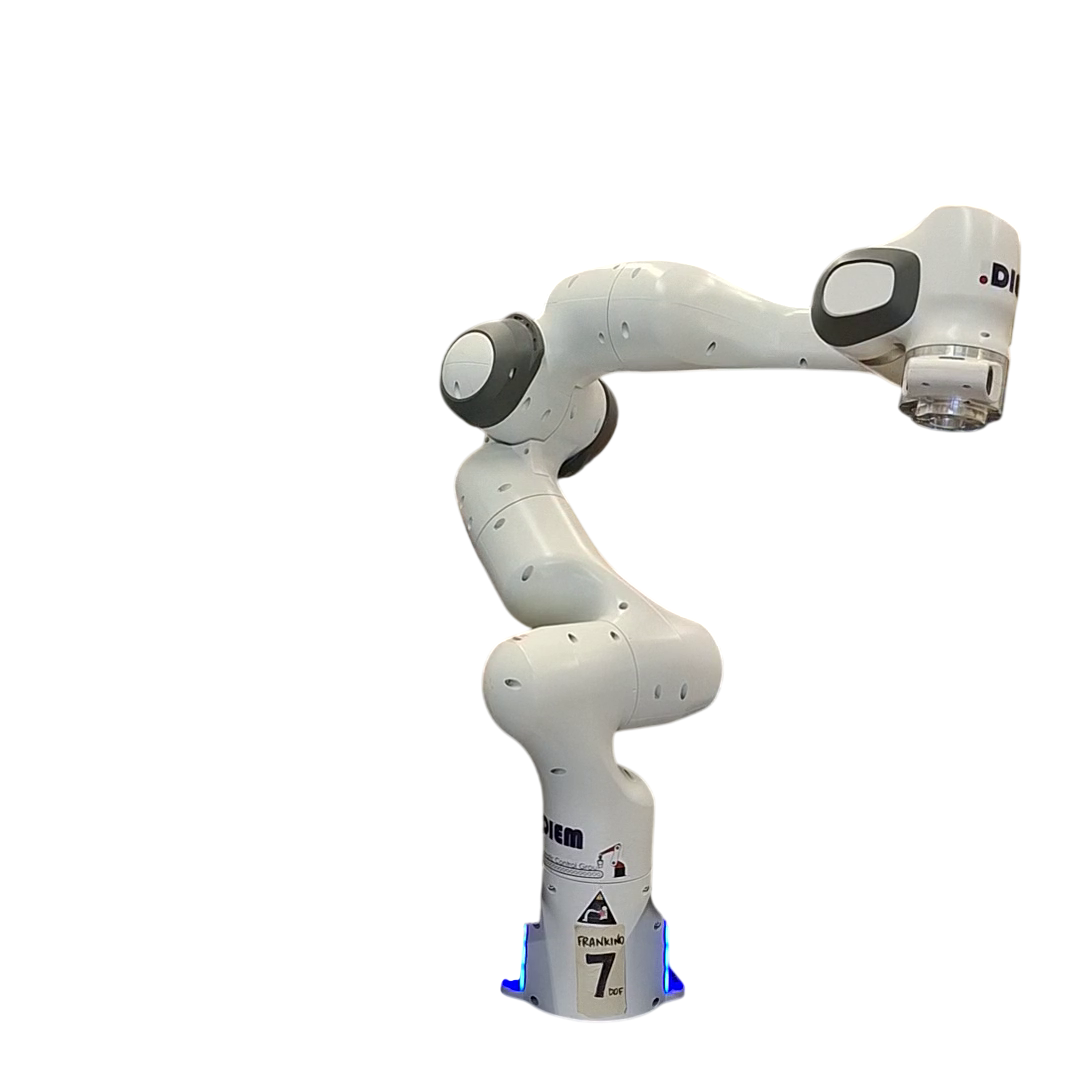}\hfill
        \includegraphics[width=0.49\linewidth]{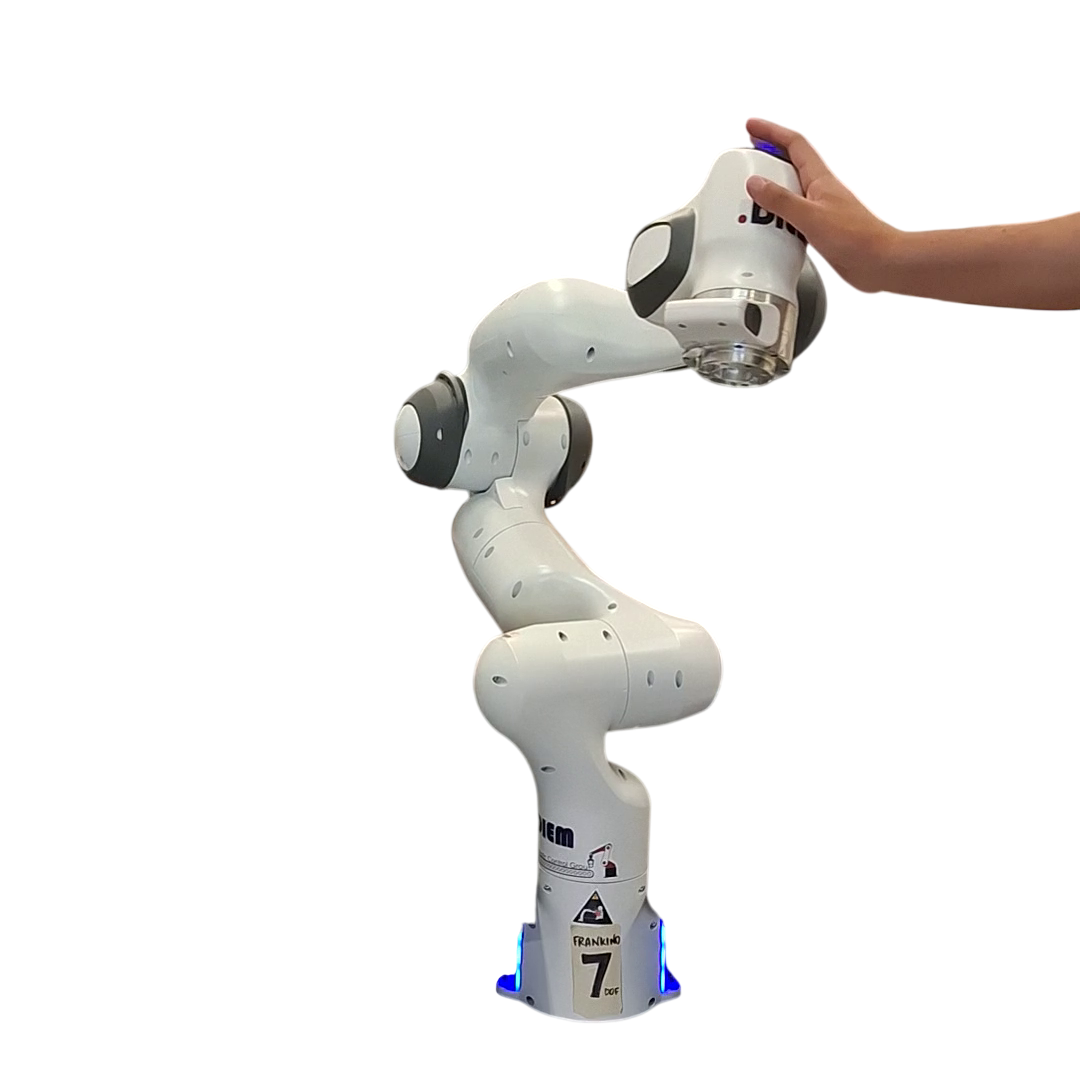}
        \caption{Salerno}
    \end{subfigure}
    \hfill
    \begin{subfigure}[b]{0.48\textwidth}
        \centering
        \includegraphics[width=0.49\linewidth]{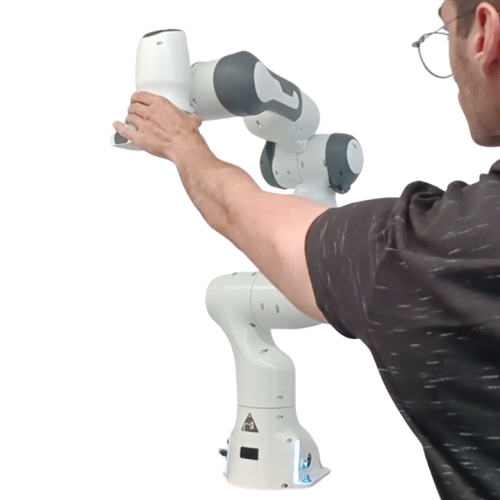}\hfill
        \includegraphics[width=0.49\linewidth]{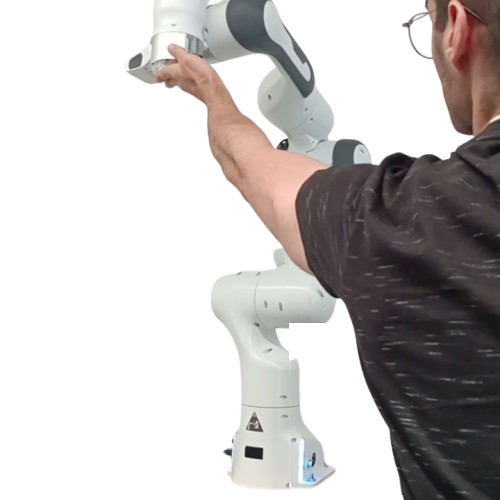}\\[1mm]
        \includegraphics[width=0.49\linewidth]{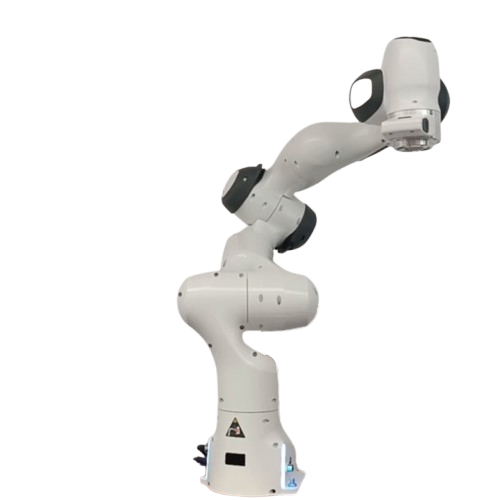}\hfill
        \includegraphics[width=0.49\linewidth]{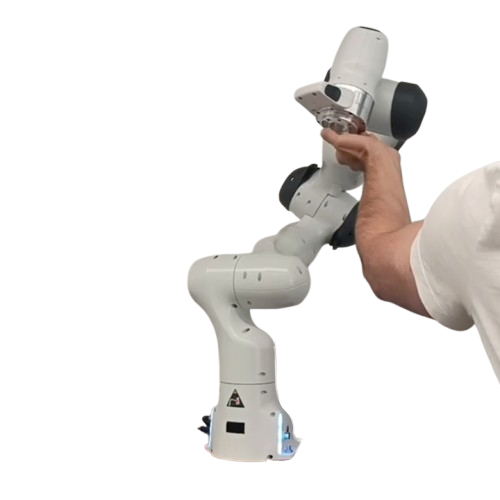}
        \caption{IDSIA}
    \end{subfigure}
    \caption{Compliance control, reproduced on both Panda setups. Within each block: (top) standstill, (bottom) trajectory tracking, (left) stiff, (right) compliant. Full videos at \url{https://sites.google.com/unisa.it/fer-ros2/applications/compliance-control}.}
    \label{fig:compliance-control}
\end{figure*}

By retuning the Cartesian gains $\mathbf K_P$ and $\mathbf K_D$ as in Table~\ref{tab:control-parameters}, the controller ranges from stiff to compliant behavior. With stiff gains, the arm holds its pose or tracks its trajectory tightly and resists contact; with compliant gains, it yields to external forces.
Fig.~\ref{fig:compliance-control} shows both regimes: at standstill (top), the robot reacts softly to a human push under compliant gains while staying firm under stiff ones; along a trajectory (bottom), the human can deflect the compliant robot, whereas the stiff run is executed without human contact for safety~(\ref{itm:experiment-compliance-control}\ref{itm:experiment-compliance-controll-trajectory-stiff}) and tracks unperturbed.
Readers are invited to see the full videos for a better visualization of these behaviors\footnote{\url{https://sites.google.com/unisa.it/fer-ros2/applications/compliance-control}}.

\subsection{Pick-and-place}
\label{sec:experiment-pick-and-place}

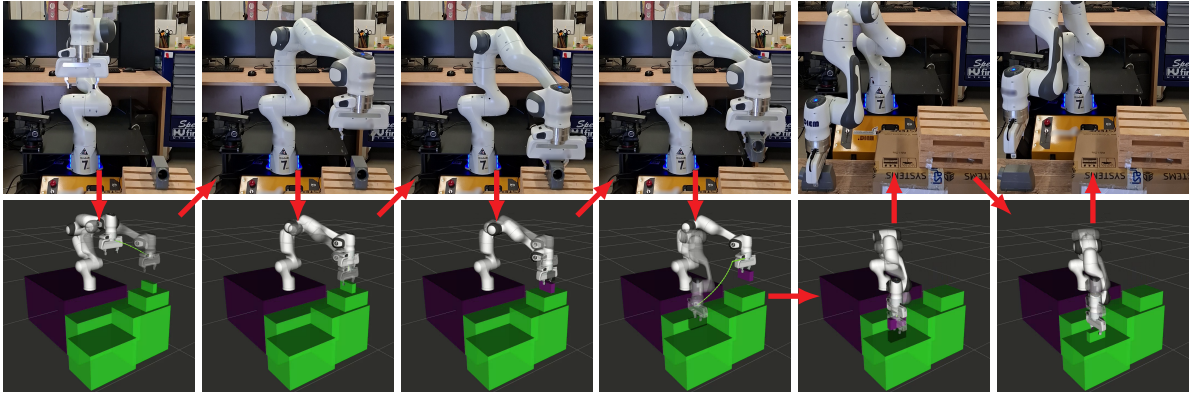
\begin{figure*}[t!]
    \centering
    \begin{tikzpicture}[
            every node/.style = {inner sep=0pt, outer sep=0pt},
            seq/.style      = {-{Latex[length=3mm]}, line width=2pt, draw=Red}
        ]
        \def\imgw{0.16\textwidth}
        \def\hgap{1mm}    
        \def\vgap{1mm}    
        \def\insetV{3mm}  
        \def\insetDx{2mm} 
        \def\insetDy{2mm} 
        \def\insetH{\insetV} 

        \node (r1c1) {\includegraphics[width=\imgw]{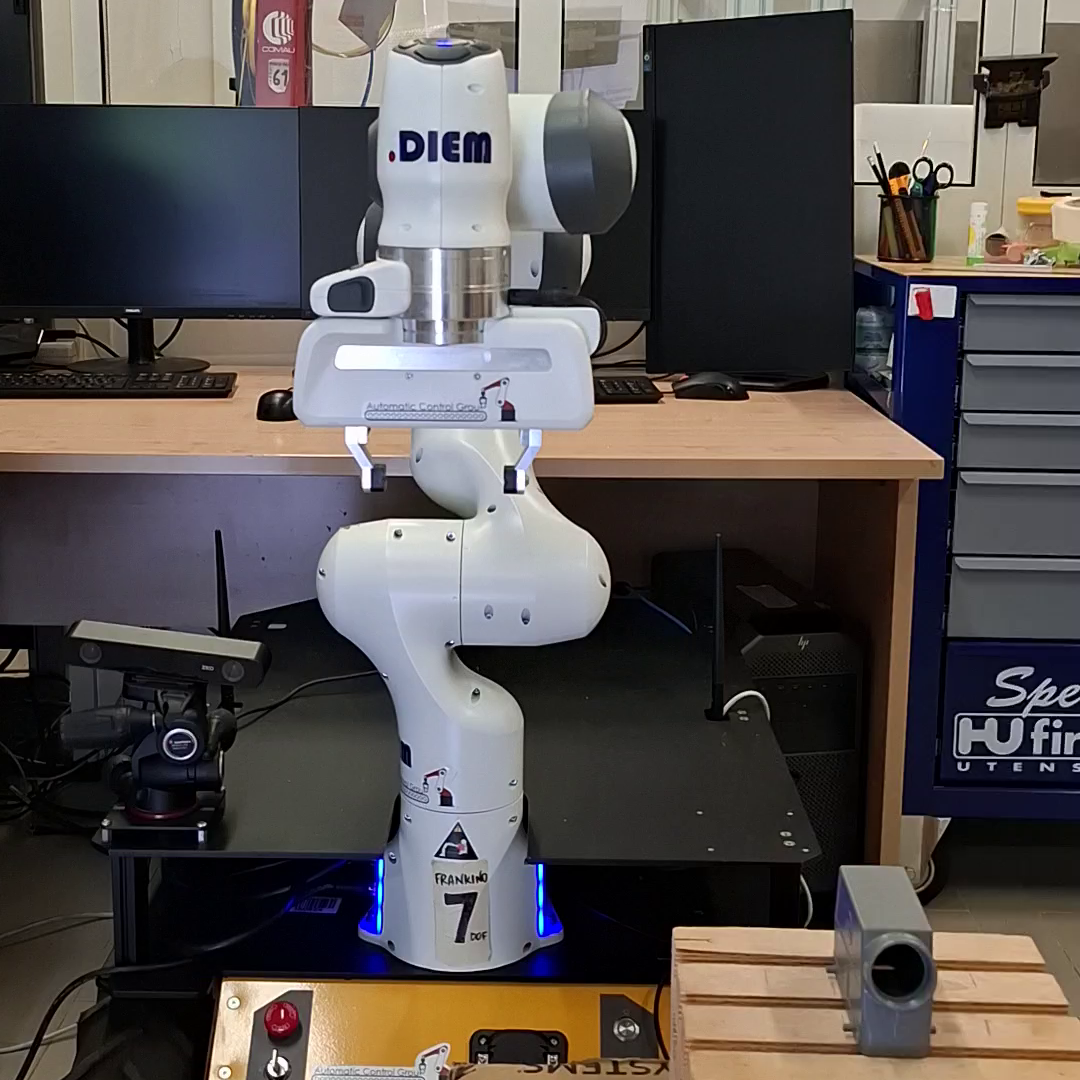}};
        \node (r1c2) [right=\hgap of r1c1] {\includegraphics[width=\imgw]{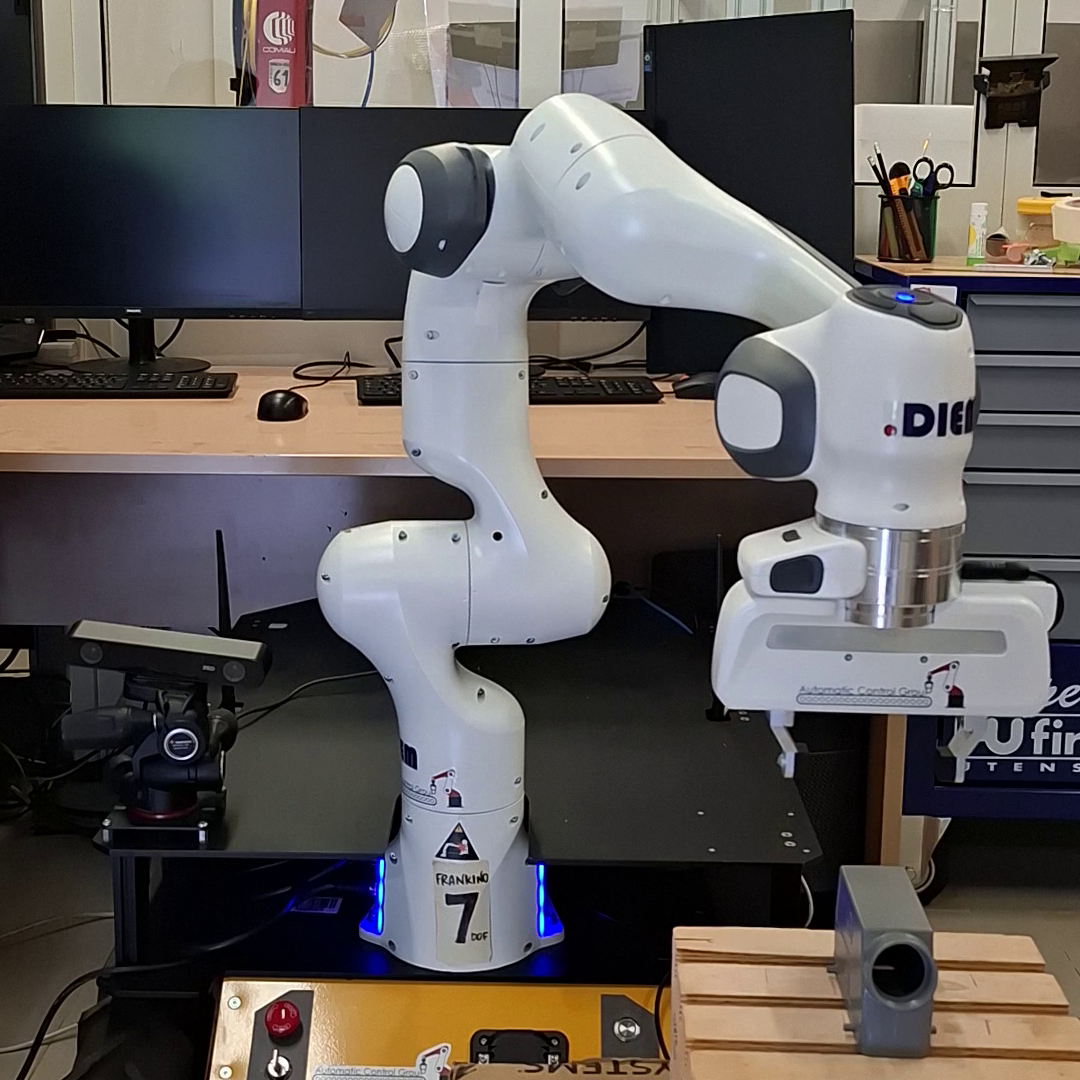}};
        \node (r1c3) [right=\hgap of r1c2] {\includegraphics[width=\imgw]{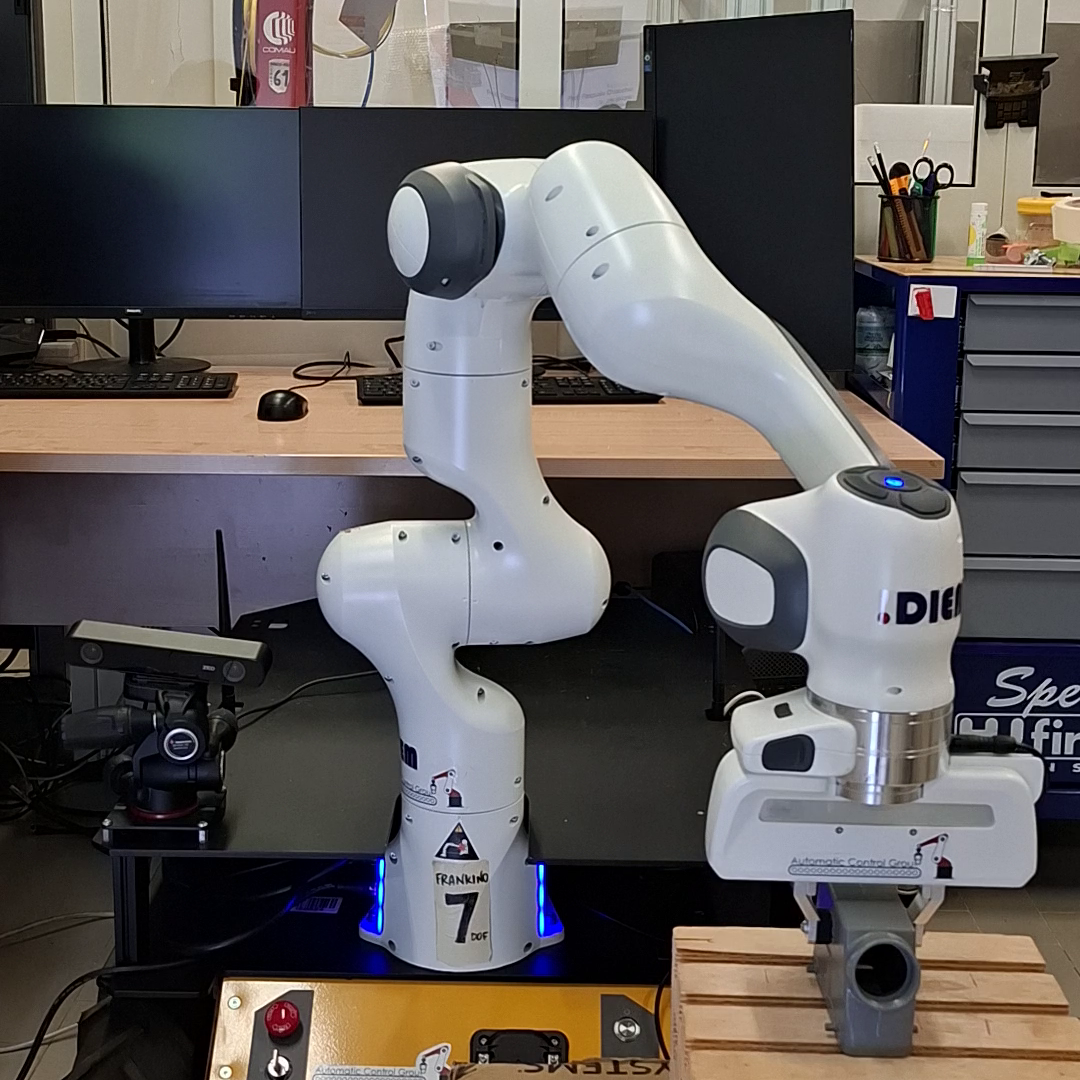}};
        \node (r1c4) [right=\hgap of r1c3] {\includegraphics[width=\imgw]{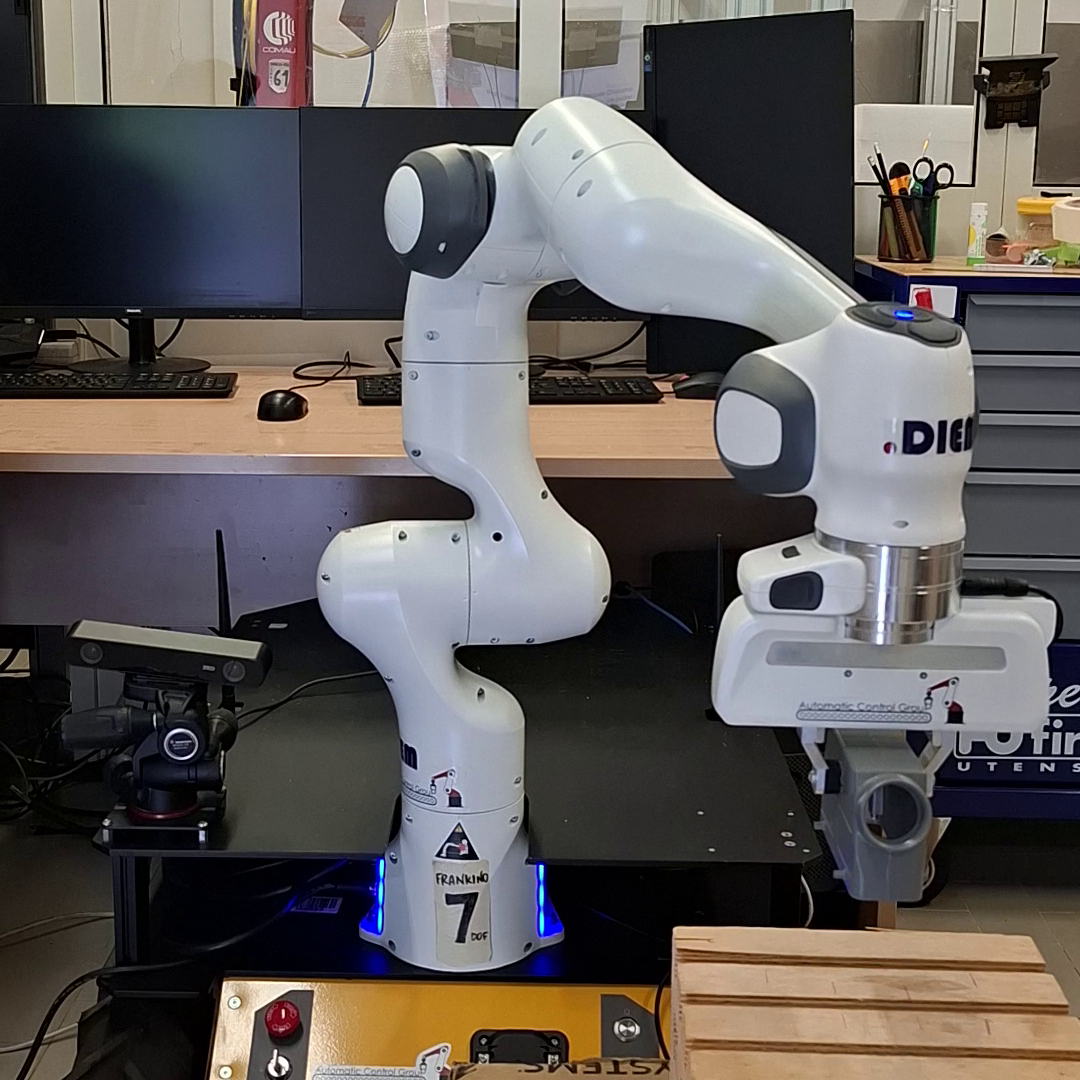}};
        \node (r1c5) [right=\hgap of r1c4] {\includegraphics[width=\imgw]{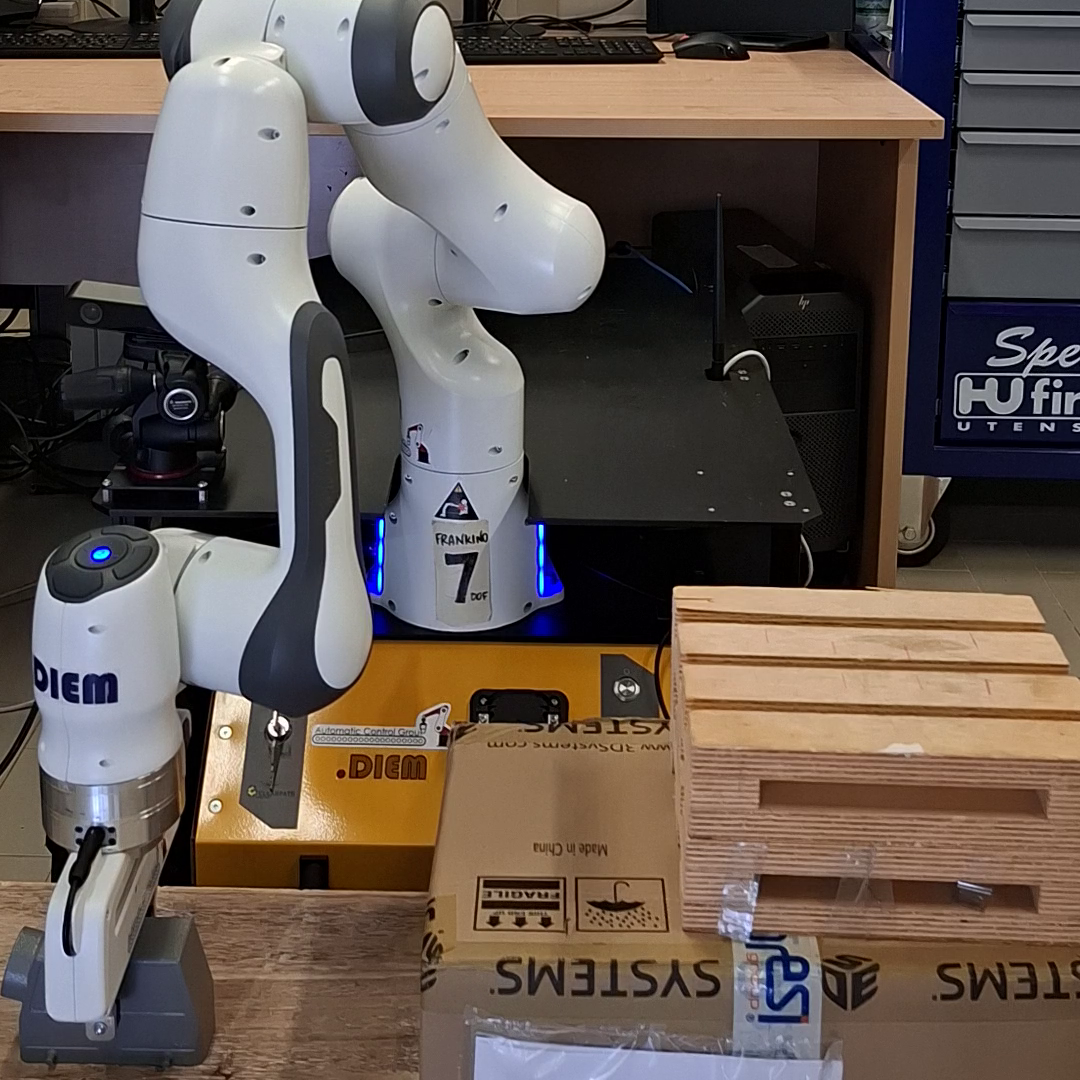}};
        \node (r1c6) [right=\hgap of r1c5] {\includegraphics[width=\imgw]{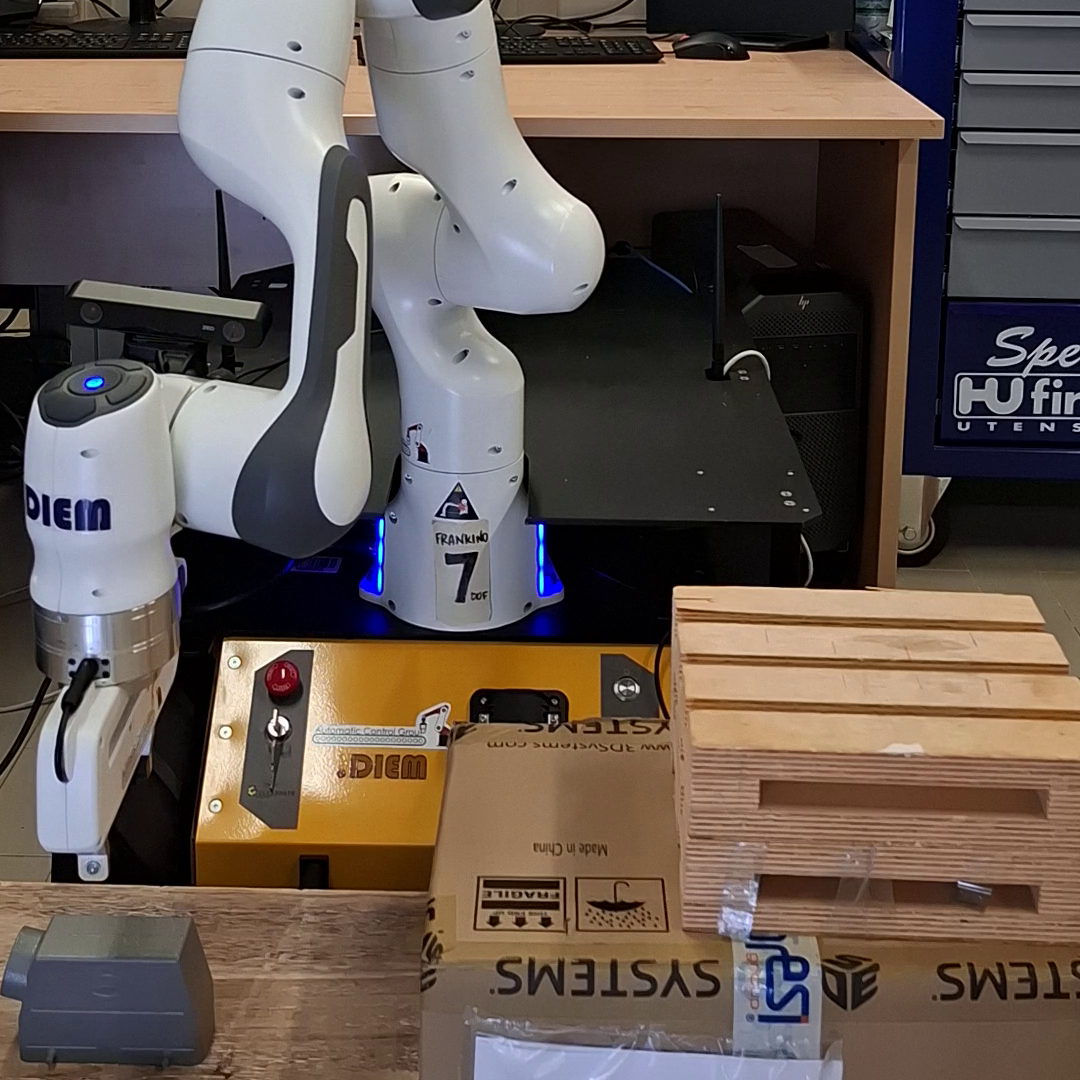}};

        \node (r2c1) [below=\vgap of r1c1] {\includegraphics[width=\imgw]{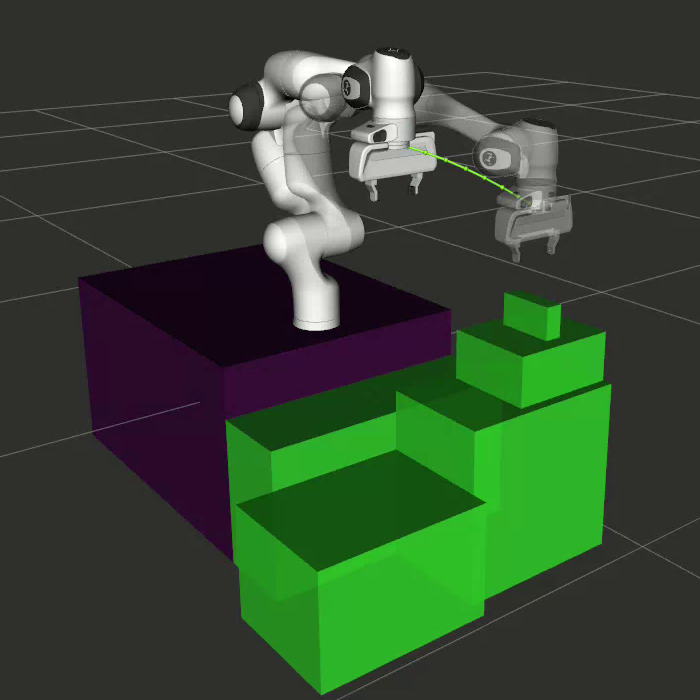}};
        \node (r2c2) [below=\vgap of r1c2] {\includegraphics[width=\imgw]{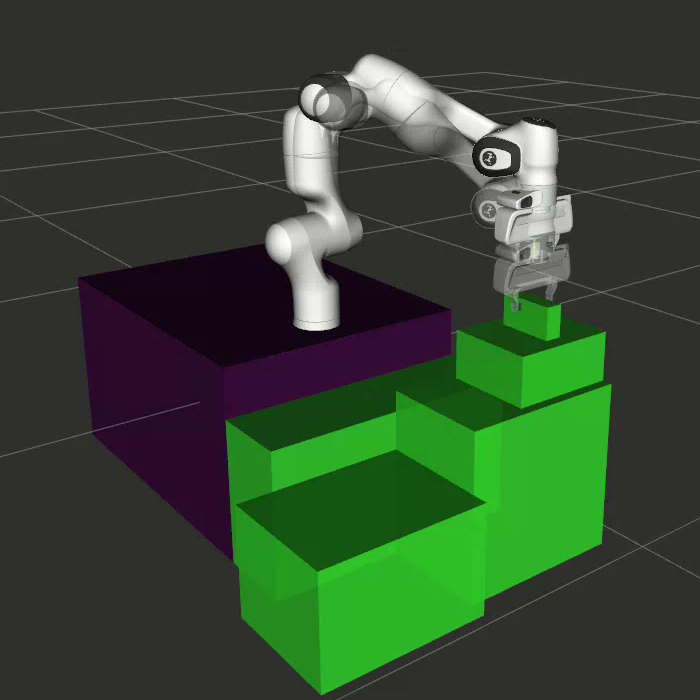}};
        \node (r2c3) [below=\vgap of r1c3] {\includegraphics[width=\imgw]{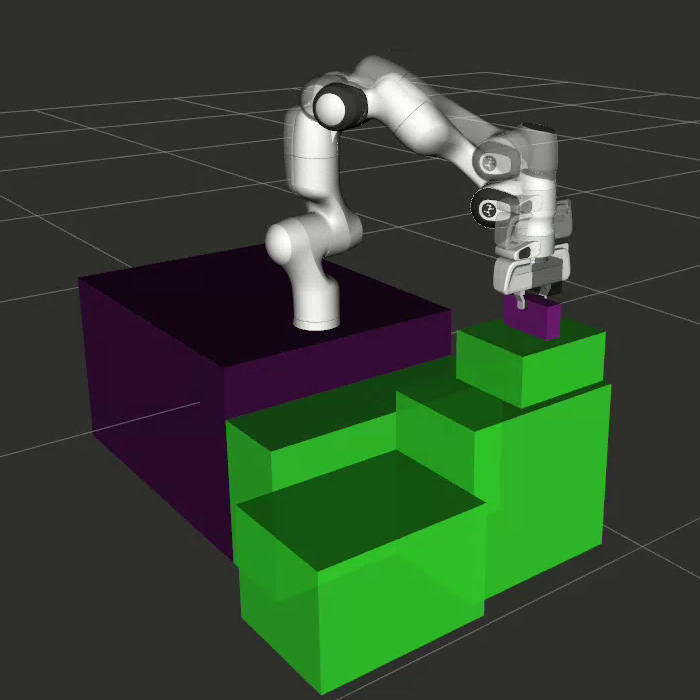}};
        \node (r2c4) [below=\vgap of r1c4] {\includegraphics[width=\imgw]{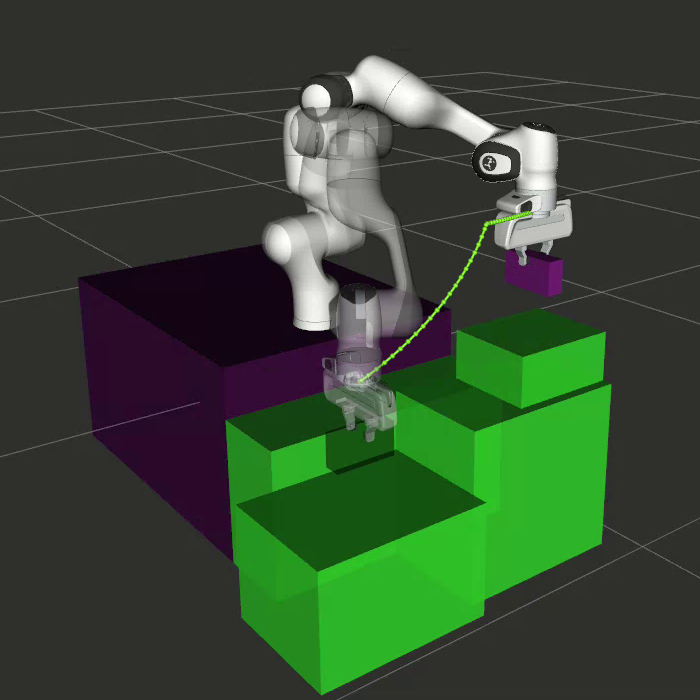}};
        \node (r2c5) [below=\vgap of r1c5] {\includegraphics[width=\imgw]{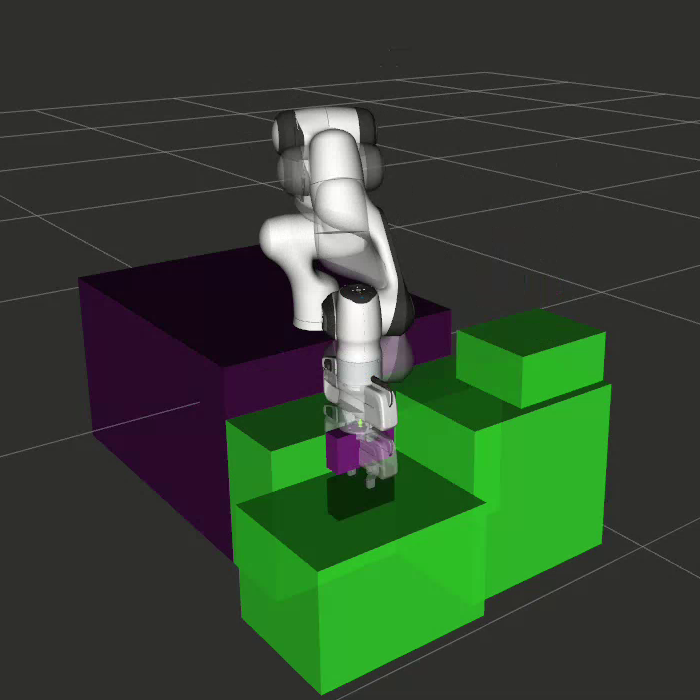}};
        \node (r2c6) [below=\vgap of r1c6] {\includegraphics[width=\imgw]{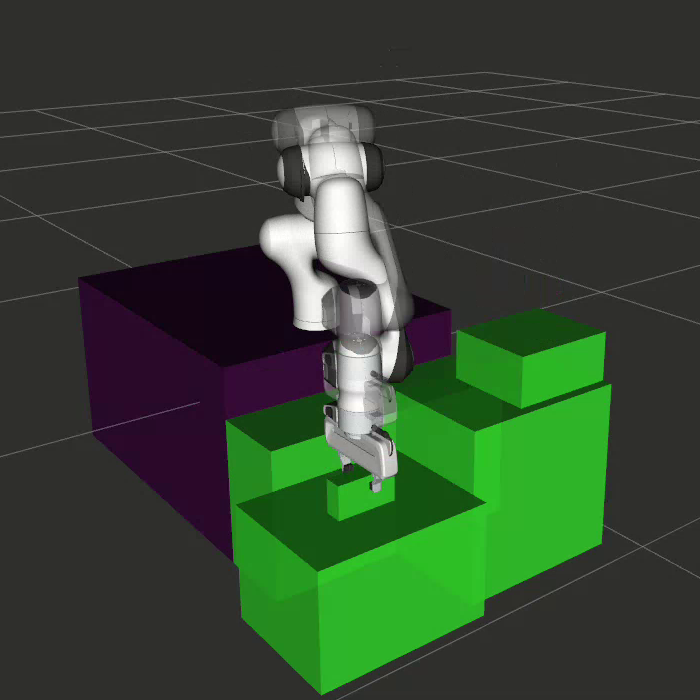}};

        \draw[seq] ($(r1c1.south) + (0,\insetV)$) -- ($(r2c1.north) + (0,-\insetV)$);
        \draw[seq] ($(r2c1.north east) + (-\insetDx,-\insetDy)$) -- ($(r1c2.south west) + (\insetDx,\insetDy)$);
        \draw[seq] ($(r1c2.south) + (0,\insetV)$) -- ($(r2c2.north) + (0,-\insetV)$);
        \draw[seq] ($(r2c2.north east) + (-\insetDx,-\insetDy)$) -- ($(r1c3.south west) + (\insetDx,\insetDy)$);
        \draw[seq] ($(r1c3.south) + (0,\insetV)$) -- ($(r2c3.north) + (0,-\insetV)$);
        \draw[seq] ($(r2c3.north east) + (-\insetDx,-\insetDy)$) -- ($(r1c4.south west) + (\insetDx,\insetDy)$);
        \draw[seq] ($(r1c4.south) + (0,\insetV)$) -- ($(r2c4.north) + (0,-\insetV)$);
        \draw[seq] ($(r2c4.east) + (-\insetH,0)$) -- ($(r2c5.west) + (\insetH,0)$);
        \draw[seq] ($(r2c5.north) + (0,-\insetV)$) -- ($(r1c5.south) + (0,\insetV)$);
        \draw[seq] ($(r1c5.south east) + (-\insetDx,\insetDy)$) -- ($(r2c6.north west) + (\insetDx,-\insetDy)$);
        \draw[seq] ($(r2c6.north) + (0,-\insetV)$) -- ($(r1c6.south) + (0,\insetV)$);
    \end{tikzpicture}
    \caption{Pick and place application on the Panda in Salerno; (top) execution, (bottom) planning with \texttt{MoveIt!}: current robot state and planned goal are shown solid and shaded, respectively. Arrows indicate the temporal sequence of operations. Full video at \url{https://sites.google.com/unisa.it/fer-ros2/applications\#h.swj8tuyvhts}.}
    \label{fig:pick-and-place}
\end{figure*}

This application runs a complete pick-and-place task through the \emph{position} command interface, the interface discouraged on the Panda by \cite{skerlj_bridging_2026} and made reliable by this work.
We adopt a second custom controller, the \textit{Cartesian pose controller}, defined by:
\begin{equation}
\begin{split}
    \dot{\mathbf q}_c&= \bigl( \mathbf J^\top(\mathbf q) \mathbf J(\mathbf q) + \alpha \mathbf I_n \bigr)^{-1} \mathbf J^\top(\mathbf q) ( \mathbf K_P \tilde{\mathbf x} + \dot{\mathbf x}_d),\\
    \mathbf q_c &= \int_t{\dot{\mathbf q}_c dt},
\end{split}
\end{equation}
with $\mathbf K_P$ set as in Table~\ref{tab:control-parameters} and $\alpha \coloneqq \num{1e-6}$.
Notably, this controller claims the \textit{position} command interfaces directly by commanding $\mathbf q_c$, thus exercising the interface this work makes reliable.

The robot grasps an object from one surface and places it, re-oriented, on another at a different height.
Fig.~\ref{fig:pick-and-place} illustrates the procedure, pairing snapshots of the real execution (top) with the \texttt{MoveIt!}\ planning requests computed through its C++ API and previewed to the user in RViz (bottom).

\subsection{Teleoperation}
\label{sec:experiment-teleoperation}

\begin{figure}
    \centering
    \begin{subfigure}[b]{0.32\columnwidth}
        \centering
        \includegraphics[width=\linewidth]{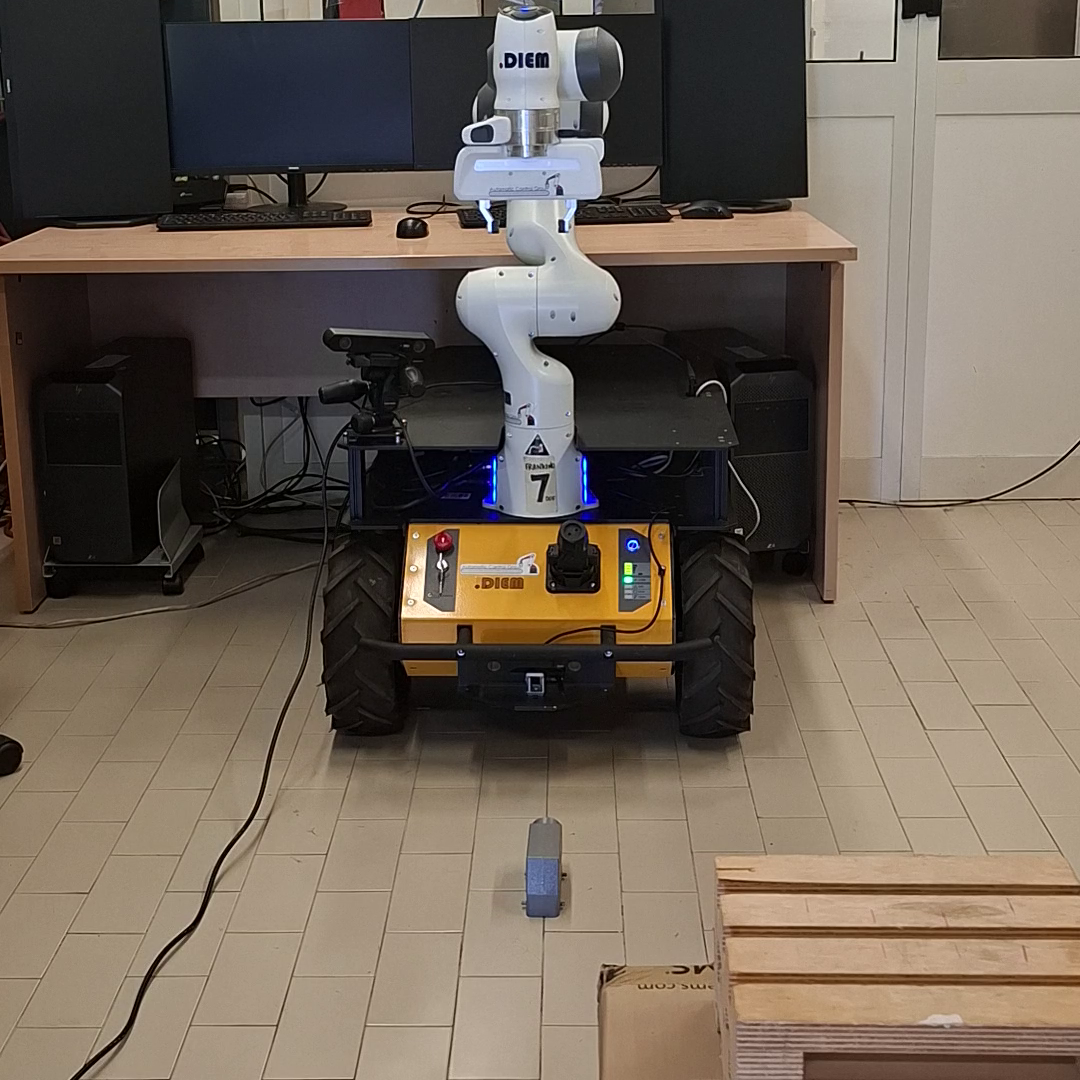}
    \end{subfigure}
    \hfill
    \begin{subfigure}[b]{0.32\columnwidth}
        \centering
        \includegraphics[width=\linewidth]{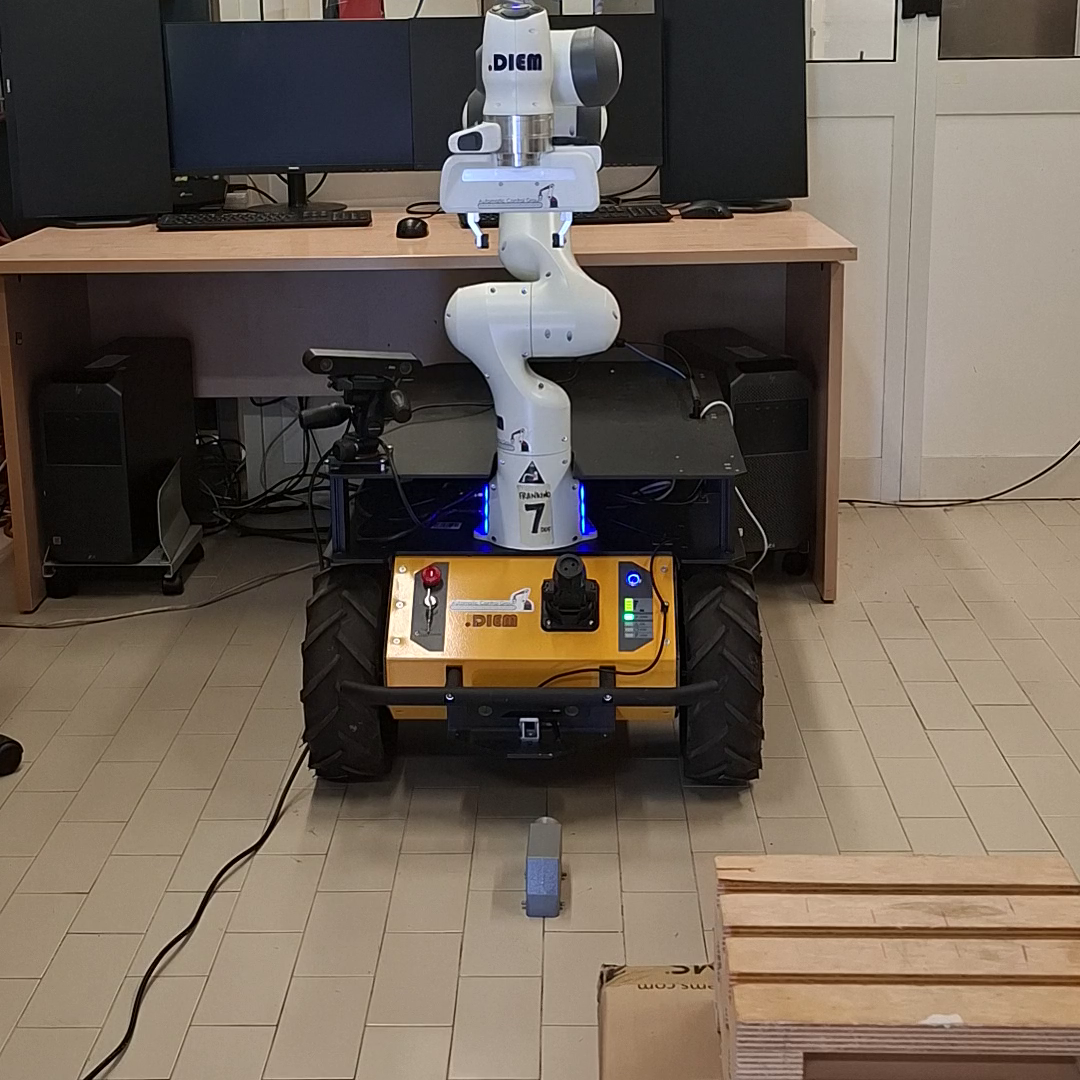}
    \end{subfigure}
    \hfill
    \begin{subfigure}[b]{0.32\columnwidth}
        \centering
        \includegraphics[width=\linewidth]{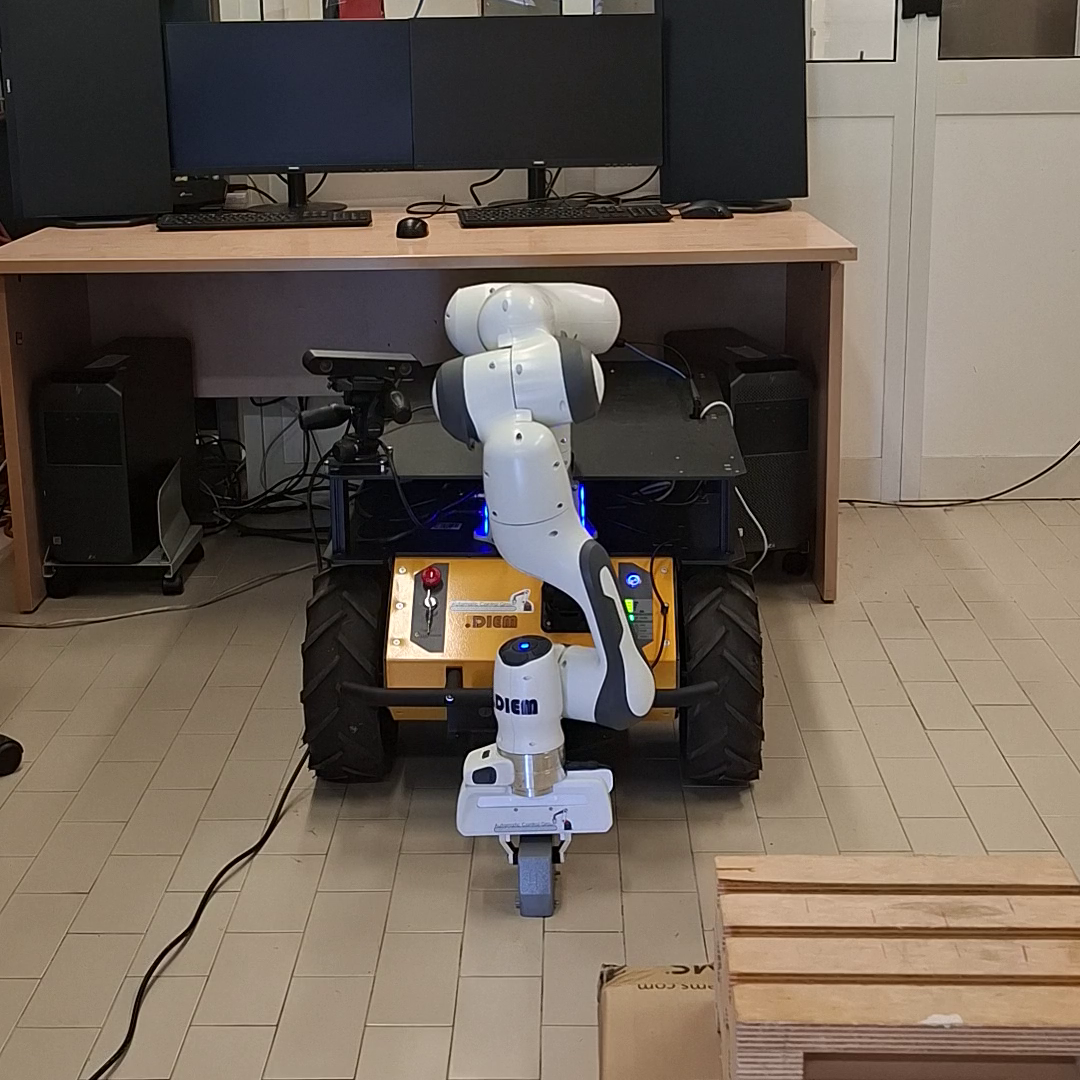}
    \end{subfigure}
    \\
    \begin{subfigure}[b]{0.32\columnwidth}
        \centering
        \includegraphics[width=\linewidth]{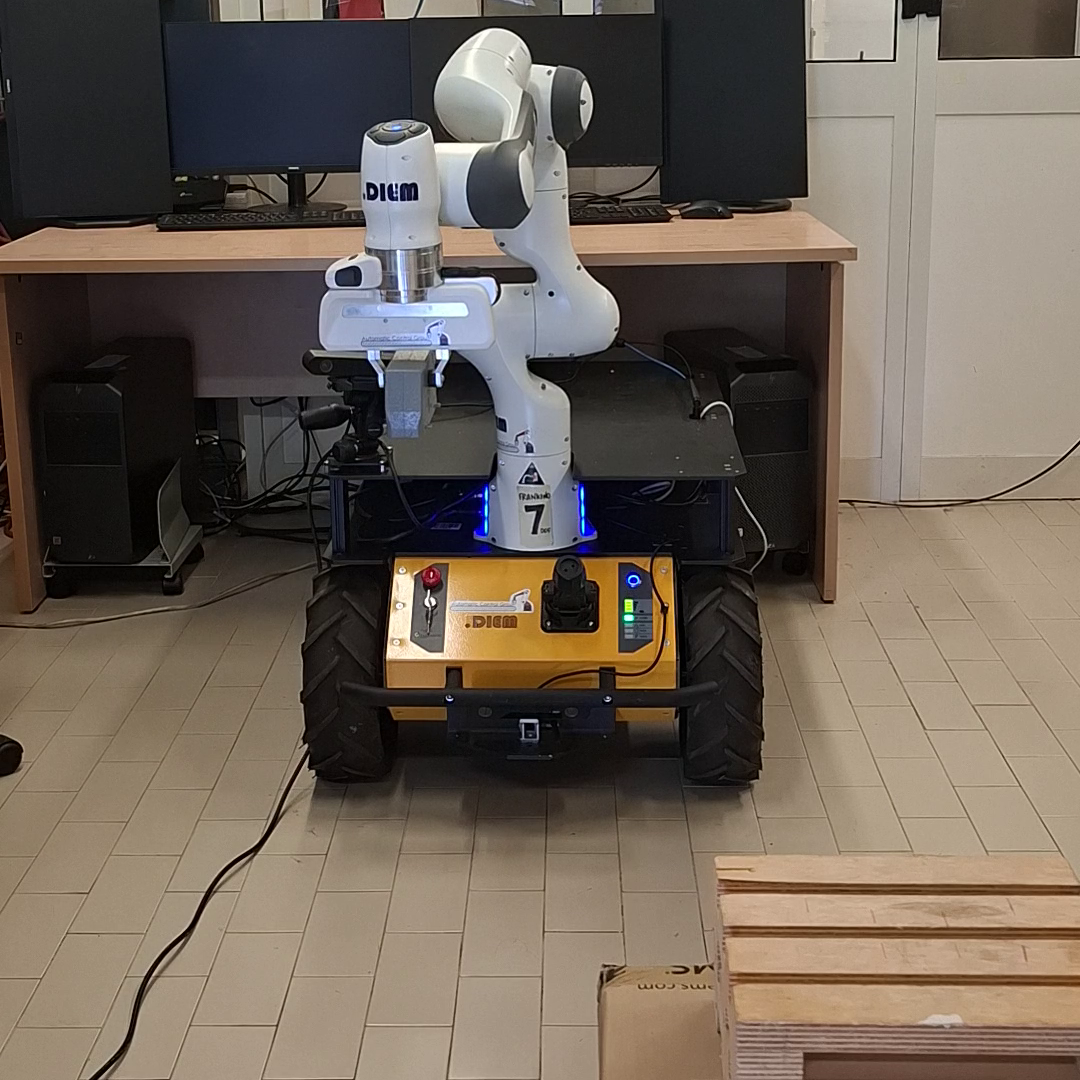}
    \end{subfigure}
    \hfill
    \begin{subfigure}[b]{0.32\columnwidth}
        \centering
        \includegraphics[width=\linewidth]{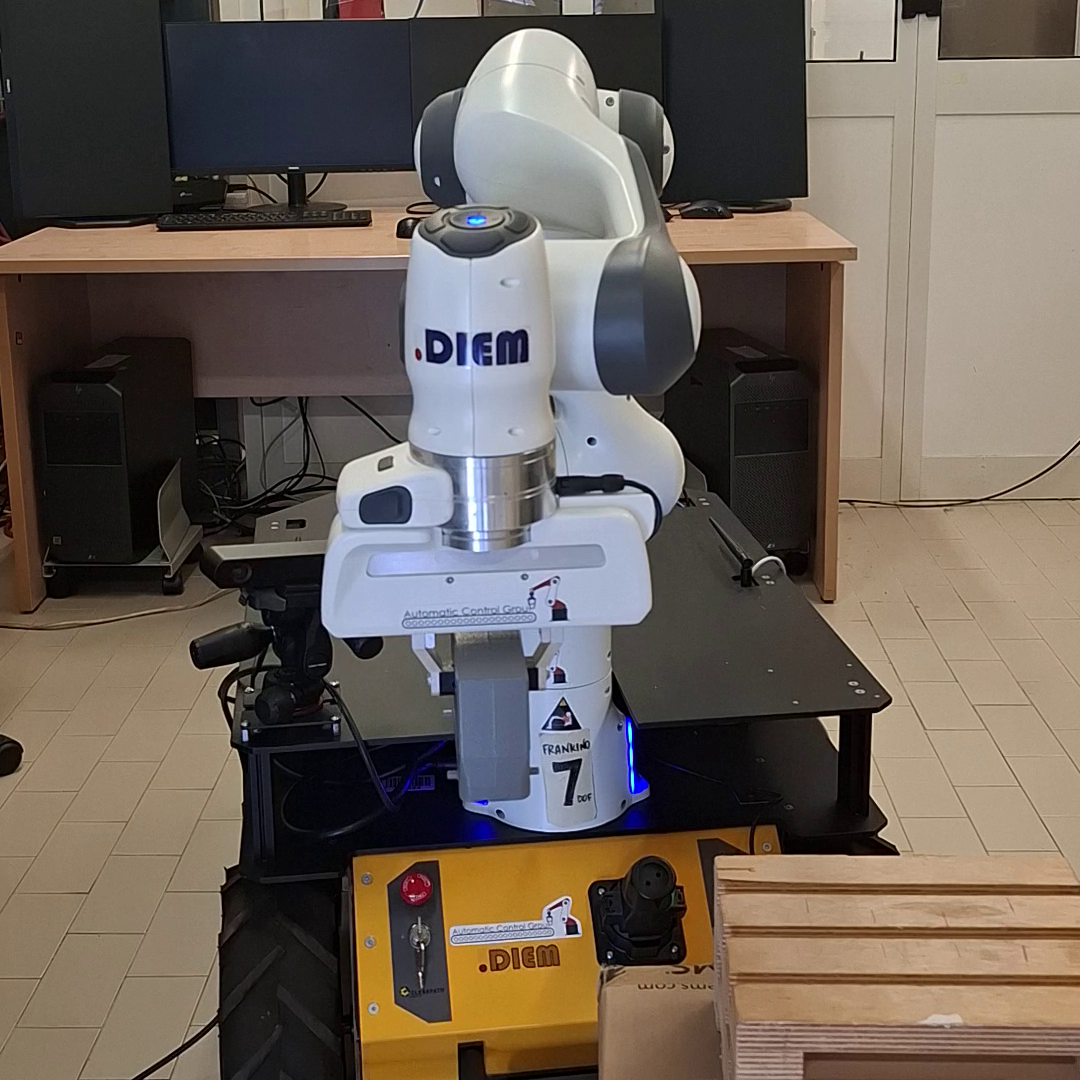}
    \end{subfigure}
    \hfill
    \begin{subfigure}[b]{0.32\columnwidth}
        \centering
        \includegraphics[width=\linewidth]{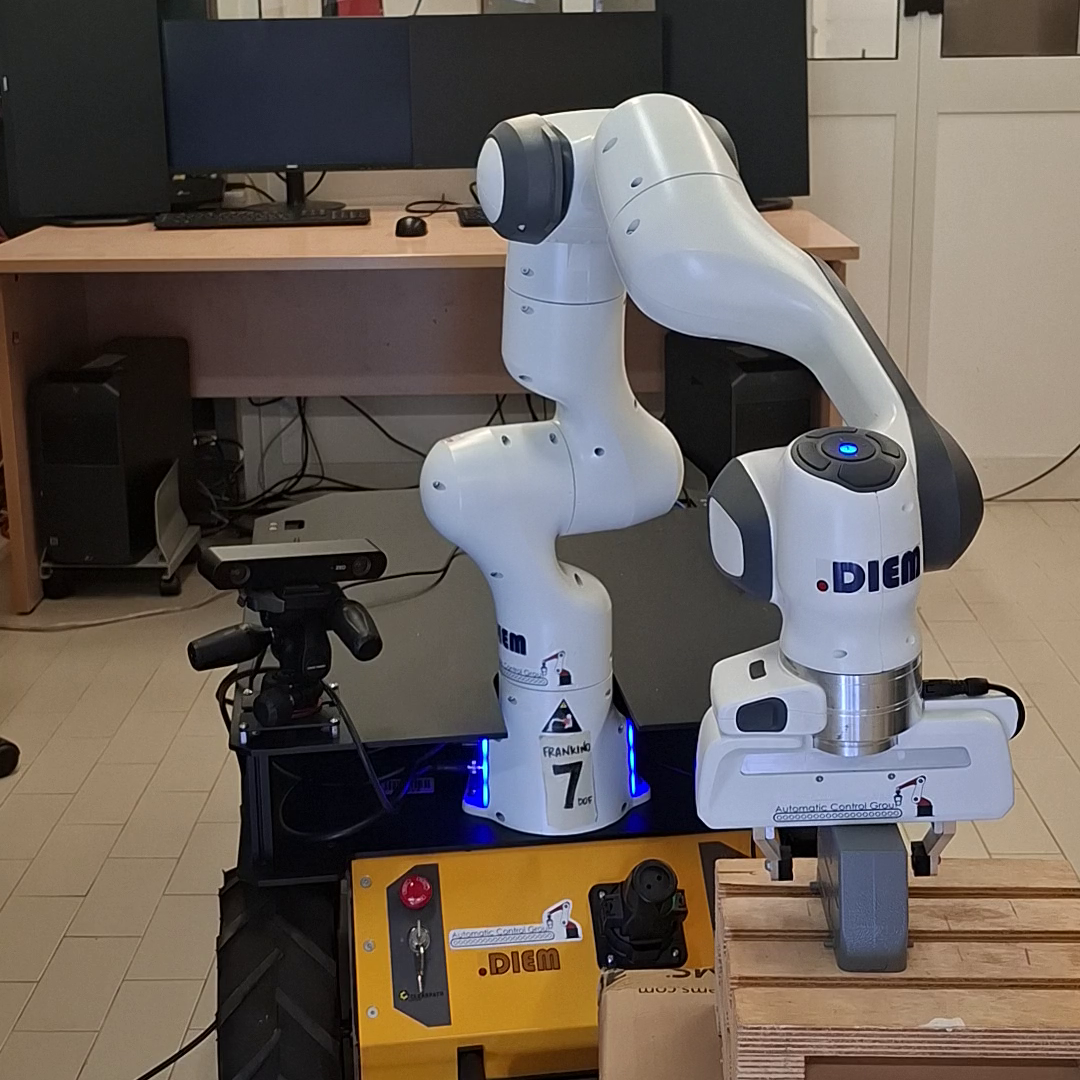}
    \end{subfigure}
    \caption{Teleoperation application in Salerno. Full video at \url{https://sites.google.com/unisa.it/fer-ros2/applications\#h.swj8tuyvhts}.}
    \label{fig:teleoperation}
\end{figure}

A human operator teleoperates the system through a \emph{haptic device}, whose motion is mapped either to Cartesian-space references ($\mathbf x_d$ in \eqref{eq:compliance-control}) for the arm or to twist commands for a Husky~A200 mobile base, alternating between the two. The arm tracks these references with the \emph{compliance controller}~\eqref{eq:compliance-control}, using the gains of Table~\ref{tab:control-parameters} and $k_0 \coloneqq 0.3$.
The task is similar to the pick-and-place of Sec.~\ref{sec:experiment-pick-and-place}, but the mobile base extends the arm's workspace: as Fig.~\ref{fig:teleoperation} shows, it carries the manipulator to objects beyond its fixed reach, which it then grasps and releases at new locations.

\section{\uppercase{Conclusions}}
\label{sec:conclusions}
 
This work brings the Franka Emika Panda back onto a current ROS~2 stack and makes its position command interface usable.
As a preliminary analysis, we traced the failure of external position control to the timing of the control loop rather than to the robot.
In the official implementation, the external workstation timer triggers the blocking FCI read, but the robot's controller dictates when new data arrives.
We concluded that, as these two clocks are independent, their relative phase drifts, causing periodic delays that leave robot cycles unserved and make the external computer distort the command stream by sampling the trajectory at instants whose jitter the onboard motion generator reads as spurious velocity and jerk.

To address this issue, the proposed architecture separates the two concerns.
A real-time thread owns the communication with the robot and keeps the external loop within the interface budget, at a cost indistinguishable from a bare \texttt{libfranka} program, while the \texttt{ros2\_control} loop is free to run at its own rate, matched to the robot's by a filtering stage.
References are generated in the position domain, anchored to the robot's own commanded state, so no clock of the workstation enters the command.
The position interface then tracks as accurately as the synchronous one while reducing the spectral content that the rate limiter leaves behind.
The four applications of Sec.~\ref{sec:applications} exercise the position, velocity and torque interfaces on two independent setups.
Code is released as open source.
 
The approach has limits that are worth stating.
The reference generation is assessed empirically: we make no formal claim on the smoothness or the stability of the resulting command stream, and the gain $\lambda$ trades the speed at which the robot's state rejoins the external reference against the size of the per-cycle correction.
The rate-matching stage costs a group delay that grows with the ratio of the two rates, which bounds how slowly the ROS side can usefully run.
Unserved cycles are reduced but not removed, since a general-purpose workstation retains a residual scheduling latency of its own.
 
Future work follows these limits.
A correction saturated at the admissible per-cycle increment would replace the fixed gain and reconcile the two states at the fastest feasible rate, which also opens the way to the constraint-aware reference generation needed under sustained packet loss.
The stack will be ported to the next ROS~2 distribution, and the position-domain scheme, which assumes nothing specific to this robot beyond the structure of its command interface, is a candidate for other manipulators whose controllers expose a commanded state.

\section*{\uppercase{Acknowledgements}}

The authors would like to thank Adamo Zito, Antonio Romano, and Federico Santonicola for their contribution in developing the software that made this work possible.
The authors warmly thank the colleagues at IDSIA, USI-SUPSI, for hosting part of the experiments conducted in this work.

\bibliographystyle{apalike}
{\small
\bibliography{bib}}

\begin{thebibliography}{}

\bibitem[Berscheid and Kröger, 2021]{berscheid_jerk-limited_2021}
Berscheid, L. and Kröger, T. (2021).
\newblock Jerk-limited {Real}-time {Trajectory} {Generation} with {Arbitrary} {Target} {States}.
\newblock In {\em Proceedings of {Robotics}: {Science} and {Systems}}.

\bibitem[Casini et~al., 2019]{casini_response-time_2019}
Casini, D., Blaß, T., Lütkebohle, I., and Brandenburg, B.~B. (2019).
\newblock Response-{Time} {Analysis} of {ROS} 2 {Processing} {Chains} {Under} {Reservation}-{Based} {Scheduling}.
\newblock In {\em 31st {Euromicro} {Conference} on {Real}-{Time} {Systems} ({ECRTS})}, volume 133 of {\em {LIPIcs}}, pages 6:1--6:23. Schloss Dagstuhl – Leibniz-Zentrum für Informatik.

\bibitem[Chitta et~al., 2017]{chitta_ros_control_2017}
Chitta, S., Marder-Eppstein, E., Meeussen, W., Pradeep, V., Tsouroukdissian, A.~R., Bohren, J., Coleman, D., Magyar, B., Raiola, G., Lüdtke, M., and Perdomo, E.~F. (2017).
\newblock ros\_control: {A} generic and simple control framework for {ROS}.
\newblock {\em The Journal of Open Source Software}, 2(20):456.

\bibitem[Daniel et~al., 2024]{daniel_multi_2024}
Daniel, M., Magassouba, A., Aranda, M., Lequièvre, L., Corrales~Ramón, J.~A., Iglesias~Rodriguez, R., and Mezouar, Y. (2024).
\newblock Multi {Actor}-{Critic} {DDPG} for {Robot} {Action} {Space} {Decomposition}: {A} {Framework} to {Control} {Large} {3D} {Deformation} of {Soft} {Linear} {Objects}.
\newblock {\em IEEE Robotics and Automation Letters}, 9(2):1318--1325.

\bibitem[de~Melo et~al., 2025]{de_melo_unified_2025}
de~Melo, V.~R., Jetti, G., Braghin, F., and Roveda, L. (2025).
\newblock Unified {Safety}-{Aware} {Physical} {Human}-{Robot} {Collaborative} {Controller} with {Online} {Continuous} {Learning} and {Personalized} {Tuning}.
\newblock In {\em 2025 {IEEE} {International} {Conference} on {Advanced} {Robotics} ({ICAR})}, pages 818--825. IEEE.

\bibitem[Featherstone and Orin, 2008]{featherstone_dynamics_2008}
Featherstone, R. and Orin, D.~E. (2008).
\newblock Dynamics.
\newblock In Siciliano, B. and Khatib, O., editors, {\em Springer {Handbook} of {Robotics}}, pages 35--65. Springer, Berlin, Heidelberg.

\bibitem[{Franka Robotics}, 2025a]{franka_robotics_franka_2025-1}
{Franka Robotics} (2025a).
\newblock Franka {Control} {Interface} {Documentation}.
\newblock \url{https://frankarobotics.github.io/docs/}.

\bibitem[{Franka Robotics}, 2025b]{franka_robotics_franka_2025}
{Franka Robotics} (2025b).
\newblock Franka {Emika} {Panda}: {Robot} and {Interface} {Specifications}.
\newblock \url{https://frankarobotics.github.io/docs/control_parameters.html}.

\bibitem[{Franka Robotics}, 2025c]{franka_robotics_franka_ros2_2025}
{Franka Robotics} (2025c).
\newblock franka\_ros2: {ROS} 2 {Integration} for {Franka} {Research} {Robots}.
\newblock \url{https://github.com/frankarobotics/franka_ros2}.

\bibitem[{Franka Robotics}, 2026]{franka_robotics_franka_ros2_2026}
{Franka Robotics} (2026).
\newblock franka\_ros2 {Releases}.
\newblock \url{https://github.com/frankarobotics/franka_ros2/releases}.

\bibitem[Haddadin et~al., 2022]{haddadin_franka_2022}
Haddadin, S., Parusel, S., Johannsmeier, L., Golz, S., Gabl, S., Walch, F., Sabaghian, M., Jähne, C., Hausperger, L., and Haddadin, S. (2022).
\newblock The {Franka} {Emika} {Robot}: {A} {Reference} {Platform} for {Robotics} {Research} and {Education}.
\newblock {\em IEEE Robotics \& Automation Magazine}, 29(2):46--64.

\bibitem[Macenski et~al., 2022]{macenski_robot_2022}
Macenski, S., Foote, T., Gerkey, B., Lalancette, C., and Woodall, W. (2022).
\newblock Robot {Operating} {System} 2: {Design}, architecture, and uses in the wild.
\newblock {\em Science Robotics}, 7(66).

\bibitem[Petrone et~al., 2026]{petrone_optimized_2026}
Petrone, V., Puricelli, L., Pozzi, A., Ferrentino, E., Chiacchio, P., Braghin, F., and Roveda, L. (2026).
\newblock Optimized {Residual} {Action} for {Interaction} {Control} {With} {Learned} {Environments}.
\newblock {\em IEEE Transactions on Control Systems Technology}, 34(2):1044--1050.

\bibitem[Reghenzani et~al., 2019]{reghenzani_real-time_2019}
Reghenzani, F., Massari, G., and Fornaciari, W. (2019).
\newblock The {Real}-{Time} {Linux} {Kernel}: {A} {Survey} on {PREEMPT}\_rt.
\newblock {\em ACM Computing Surveys}, 52(1):18.

\bibitem[Risi et~al., 2026]{risi_simplifying_2026}
Risi, D., Petrone, V., Langella, A., Pagliara, L., Ferrentino, E., and Chiacchio, P. (2026).
\newblock Simplifying {ROS2} controllers with a modular architecture for robot-agnostic reference generation.
\newblock {\em IEEE Robotics and Automation Practice}, pages 1--5.

\bibitem[Shahid et~al., 2025]{shahid_gear_2025}
Shahid, A.~A., Moroncelli, A., Brscic, D., Kanda, T., and Roveda, L. (2025).
\newblock {GEAR}: {Gaze}-{Enabled} {Human}-{Robot} {Collaborative} {Assembly}.
\newblock In {\em 2025 {IEEE}/{RSJ} {International} {Conference} on {Intelligent} {Robots} and {Systems} ({IROS})}, pages 4457--4464. IEEE.

\bibitem[Tingelstad, 2024]{tingelstad_libfranka_2024}
Tingelstad, L. (2024).
\newblock libfranka with {Active} {Control} for the {Franka} {Emika} {Panda}.
\newblock \url{https://github.com/tingelst/libfranka}.

\bibitem[Todorov et~al., 2012]{todorov_mujoco_2012}
Todorov, E., Erez, T., and Tassa, Y. (2012).
\newblock {MuJoCo}: {A} physics engine for model-based control.
\newblock In {\em 2012 {IEEE}/{RSJ} {International} {Conference} on {Intelligent} {Robots} and {Systems}}, pages 5026--5033. IEEE.

\bibitem[Škerlj et~al., 2026a]{skerlj_bridging_2026}
Škerlj, J., Bien, S., Naceri, A., and Haddadin, S. (2026a).
\newblock Bridging the {Sim}-to-{Real} {Gap} with multipanda\_ros2: {A} {Real}-{Time} {ROS2} {Framework} for {Multimanual} {Systems}.
\newblock \url{https://doi.org/10.48550/arXiv.2602.02269}.
\newblock arXiv preprint.

\bibitem[Škerlj et~al., 2026b]{skerlj_multipanda_ros2_2026}
Škerlj, J., Bien, S., Naceri, A., and Haddadin, S. (2026b).
\newblock multipanda\_ros2: {A} {Sim}- and {Real} {Panda} {Robot} {Integration} {Based} on the ros2\_control {Framework}.
\newblock \url{https://github.com/tenfoldpaper/multipanda_ros2}.

\end{thebibliography}

\end{document}